\documentclass[onecondense,final]{svjour3}       
\usepackage{natbib}
\usepackage{url}
\usepackage{multicol}
\usepackage{setspace}
\usepackage{ulem}

\begin{document}

\title{An Account of Opinion Implicatures}

\author{Janyce Wiebe         \and
        Lingjia Deng
}

\institute{Janyce Wiebe\at
              Department of Computer Science \\
              University of Pittsburgh\\
              \email{wiebe@cs.pitt.edu}           
           \and
           Lingjia Deng \at
              Intelligent Systems Program \\
              University of Pittsburgh \\
              \email{lid29@pitt.edu}
}

\maketitle

\begin{abstract}
While previous sentiment analysis research has concentrated on the
interpretation of explicitly stated opinions and attitudes, this work
initiates the computational study of a type of opinion implicature
(i.e., opinion-oriented inference) in text.   This paper
described a rule-based framework for representing and analyzing
opinion implicatures which we hope will contribute to deeper automatic
interpretation of subjective language.  In the course of understanding
implicatures,
the system recognizes implicit sentiments (and beliefs)
toward various events and entities in the sentence, often attributed
to different sources (holders) and of mixed
polarities; thus, it produces a richer interpretation than is
typical in opinion analysis.
\end{abstract}

\section{Introduction and Motivation}
\label{intro}

Opinions are ubiquitous in language: they appear in written and spoken
text ranging from editorials, reviews, blogs, and
informal conversations to written news reports and
broadcast news.
We have seen a dramatic surge in \textit{opinion analysis} (\textit{sentiment analysis}) research over the past decade
(\cite{bingliu2012,bobook}).
And many advances have been made with respect to the automatic identification
and characterization of opinions in text.
At the document level, methods now exist to determine whether the
overall sentiment in a review or otherwise opinion-oriented text is
positive or negative (e.g., \cite{das01}, \cite{bo02}, \cite{dave03}) and whether a sentence is
subjective or objective (e.g., \cite{wiebe:riloff:cicling:2005}).  There has also been work
on fine-grained opinion analysis: to recognize expressions of opinion
at the phrase level (e.g., \cite{eric07}, \cite{choi06}); to determine their polarity
(e.g., \cite{wilsonetal2005b}, \cite{choi-cardie:2008:EMNLP}); to identify the
opinion holder (e.g., \cite{choi05}, \cite{kim05,choi06}) --- the person or entity that is
the source of the opinion; and to identify the target or topic of the opinion
(e.g., \cite{Qiu:2011:OWE:1970420.1970422}, \cite{yietal03}, \cite{sto08}).

Still, research in opinion analysis has plateaued at a somewhat
superficial level, providing methods that exhibit a fairly shallow
understanding of subjective language as a whole.
In particular, past research in NLP has mainly addressed explicit
opinion expressions, ignoring implicit opinions expressed via
implicatures, i.e., default inferences.
Consider, for example, the following sentences:
\begin{quote}
(1) John was glad that Mary saved Bill.\\
(2) The international community seems to be tolerating the Israeli
    campaign of suppression against the Palestinians.
\end{quote}
\noindent
While existing opinion analysis techniques might be able to determine
the surface opinions (e.g., John has a positive attitude toward Mary's saving Bill),
they would stop short of identifying the sentences' many \textit{opinion
implicatures}, i.e., implied attitudes and opinions, such as the following:

\small
\begin{description}
   \item [for (1):] {\it John is positive
toward Bill}, {\it John is positive towards Mary} (for saving Bill),
{\it John believes that Mary is positive towards Bill} (because she
saved him).
   \item [for (2):] {\it The writer is negative toward the Israeli
   campaign of suppression and toward the International Community}
   (for tolerating it).
\end{description}
\normalsize

Overall, the goal of this work is to make progress toward a deeper
automatic interpretation of opinionated language by developing
computational models for the representation and interpretation of
opinion implicature in language.
This report focuses on a rule-based
implementation of a conceptual framework for opinion implicature,
specifically implicatures that arise in the presence of explicit
sentiments, and events that positively or negatively affect entities.
To eliminate interference
introduced by the noisy output of automatic NLP components, the system
takes as input manually annotated explicit-sentiment and event
information, and makes inferences based on that input information.
Thus, the purpose of this work is to provide a conceptual
understanding of (a type of) opinion implicature, to provide a
blueprint for realizing fully automatic systems in the future.

To make the framework more accessible, we begin this report with
overview sections, first motivating the framework from an NLP
perspective in Section \ref{nlp}, then providing terminology and
sketching out its processing in Section \ref{overview}.

The graphical data structure representing the system's knowledge is
described in Section
\ref{datastructure}.

The knowledge representation scheme for the
nodes and edges of the graph is presented in Section \ref{kr}; {\bf this
section may be safely skipped} for readers not concerned with such details.

The inference mechanisms are described next, in Section
\ref{inferencesGeneralDescription}, namely specifications of (1) the default
inference rules used by the system (Subsection \ref{rulespec}), (2)
the mechanism for drawing inferences in belief and sentiment spaces
(Subsection \ref{psspaces}), and (3) the cases when inferences are
blocked (Subsection \ref{blocking}).

The semantic compositions performed by the system are described in
Section \ref{compositionality} and, at last, the actual implicature rules are
given in Section \ref{therules}.  The control of execution is
described in Section \ref{control}.

Section \ref{examples} is the heart of the report, in that it steps
through examples illustrating the system's lines of reasoning.
{\bf Some readers may prefer to jump from the overview sections
  directly to
this section}, referring to the intervening sections
according to their interests.  Section \ref{exploring} revisits an
earlier example, now that the reader is familiar with the various
inference patterns, to illustrate interdependent ambiguities.

Sections \ref{negBelMeaning} and \ref{extensionsThatCouldBeAdded} return to
issues and possible extensions of the knowledge representation scheme and
belief and sentiment space mechanisms, respectively.  {\bf These may be
safely skipped}.

Section \ref{relatedwork} discusses related work.  We first consider
recent work in NLP on sentiment analysis from the perspective of our framework,
and then acknowledge older work in NLP and AI whose ideas we
exploited to create the inference architecture.
Finally, Section \ref{conclusions} is the conclusion.
Appendix A 
gives the full output of the system for the examples
covered in Sections \ref{sentobject} through \ref{evidenceAgainstEGs}.

{\bf Appendix A is available at\\
 http://www.cs.pitt.edu/$\scriptsize{\sim}$wiebe/AppendixA.txt}

\section{NLP Perspective}
\label{nlp}

As mentioned above, our rule-based system was developed to provide a
conceptual understanding of a type of opinion implicature.  Between
the rule schemas and the mechanisms for inference within sentiment and
belief spaces, it produces rich interpretations, as will be seen below
in Section \ref{examples}.  However, the system is currently supplied
with manual annotations of opinion and event information, and its
rule-based architecture is not ideal for practical application to
real-world texts.  So, before diving into presenting the
rule-based system, we pause and consider opinion implicature from the
perspective of NLP.

Rather than the explicit application of inference rules, we believe
{\it sentiment propagation} will be a key mechanism in practice.\\

\noindent
Consider the following sentence:\\

\noindent
{\bf The bill would lower health care costs, which would be a tremendous positive change across the entire health-care system.} \\

\noindent
The writer is clearly positive toward the idea of \textit{lowering
health care costs}.  But how does s/he feel about the costs?  If s/he
is {\bf positive} toward the idea of \textit{lowering} them, then,
presumably, she is {\bf negative} toward the \textit{costs} themselves
(specifically, how high they are).  The only explicit sentiment
expression, \textit{tremendous positive change}, is positive, yet we
can infer a {\bf negative} attitude toward the {\bf object} of the
event itself (i.e., \textit{health care costs}).

Going further, since \textit{the bill} is the \textbf{agent} of an
event toward which the writer is positive, we may (defeasibly) infer
that the writer is {\bf positive} toward {\it the bill}, even though
there are no explicit sentiment expressions describing it.

Now, consider \textit{The bill would curb skyrocketing health care
costs.}  The writer expresses an explicit \textbf{negative} sentiment
(\textit{skyrocketing}) toward the {\bf object} (\textit{health care
costs}) of the event.  Note that \textit{curbing} costs, like
\textit{lowering} them, is bad for them (the costs are reduced).  We
can reason that, because the event is bad for something toward which
the writer is negative, the writer is {\bf positive} toward the {\bf
event}.  We can reason from there, as above, that the writer is {\bf
positive} toward \textit{the bill}, since it is the {\bf agent} of the
positive event.

These examples illustrate how explicit sentiments toward one entity
may be propagated to other entities via opinion implicature rules.
In \cite{dengwiebeeacl2014}, we incorporate constraints derived from
implicature rules into a graph-based model, and use Loopy Belief
Propagation (\cite{Pearl1982}) to accomplish sentiment propagation in
the graph.   We showed that the graph-based model improves over an
explicit sentiment classification system.

A fully automatic implicature system will require several
computational components, each tackling its own ambiguities, such as
explicit sentiment recognition, event extraction, and semantic role
labeling.  This raises opportunities for interdependent ambiguity
resolution.  The implicature rules model dependencies among
ambiguities, such that the total number of joint interpretations is
greatly reduced.  Thus, rather than having to take a pipeline
approach, an optimization framework may exploit those
interdependencies to jointly resolve ambiguities.  In a first
study,\footnote{Currently in submission.}  we develop local
classifiers to resolve four individual ambiguities, and then use
Integer Linear Programming to conduct global inference, resulting in
substantial improvements for two of the ambiguities without loss in
performance for the others.

The studies so far only address sentiments of the writer, and they
only exploit simplified versions of four out of ten rule schemas
currently incorporated into the rule-based system.  The rule-based
system is meant to be a ``what-if'' system that looks beyond the
current capabilities of fully automatic systems toward deeper
interpretations that would be possible with improved NLP tools; it
provides an understanding that should be helpful in designing future
experiments in sentiment analysis.

\section{Overview}
\label{overview}

This section gives an overview, starting with the main concepts and
terminology underlying this work
(in Subsection \ref{background}), then introducing the system's opinion
inferences in Subsection \ref{overview:inference}.

\subsection{Some Terminology}
\label{background}

The fundamental building blocks of our opinion implicature framework
are {\it subjectivity}, {\it inferred private states}, and {\it
benefactive/malefactive}
events and states.

\subsubsection{Subjectivity}
In our work (\cite{mpqa,wiebe1994}),
{\it subjectivity}
is defined as
the expression of private states in language,
where private states are mental and emotional states such as
speculations, evaluations, sentiments, and beliefs (\cite{quirketal1985}).
We focus on three main types of subjective expressions:

\begin{description}
   \item {References to \textit{private states}:}\\
     (1) Japan has been
     {\bf eager} for a sign that Mr. Bush is {\bf concerned} about the
     issue.

   \item {References to \textit{speech or writing} events expressing private states:}\\
(2) UCC/Disciples leaders {\bf roundly condemned} the Iranian President's
{\bf verbal assault} on Israel.

   \item {\textit{Expressive subjective elements}}, which do not {\it refer} to
     private states, but rather evoke them through wording and
     description:\\
     (3) The {\bf ill-conceived} plan is based on {\bf little more} than continuation of a {\bf business-as-usual} path.
\end{description}

Subjective expressions 
have \textit{sources} (or \textit{holders}): the entity or entities whose
private states are being expressed.  For example, in (1) the source of
the private state \textit{eager} is Japan and the source of \textit{concerned}
is Mr. Bush.  In (2), the source of the subjective expressions is
\textit{UCC/Disciples leaders} and in (3) it is the writer.
Sources are, in a sense, nested: 
for example, in (1), the source of \textit{eager} is (writer, Japan), i.e.,
it is according to the writer that Japan is eager; and the source of
\textit{concerned} is (writer, Japan, Mr. Bush).

Thus, in our approach, a {\bf private state} is an attitude held by a source
toward (optionally) a target.  {\bf Sentiment and belief are types of attitudes}.
Subjectivity is the linguistic expression of private states.  {\bf Subjectivity is a
pragmatic notion}: as the sentence is interpreted in context, a private
state {\bf is} attributed to a source in that context
(\cite{banfield1982}).
By {\bf sentiment expression} or {\bf explicit sentiment}, we mean a subjective expression
where the attitude type of the expressed private state is sentiment.
We use {\bf opinion} when a general/vague term is appropriate.

There are many types of linguistic clues that contribute to
recognizing subjective expressions (\cite{wiebe1994}).
In the clearest case, some word
{\bf senses} give rise to subjectivity whenever they are used in
discourse (\cite{wiebe-mihalcea:2006:COLACL}), for example the \textit{hindrance} meaning of
\textit{catch} (\textit{what's the catch?}).  Other clues are not as definitive.
For example, researchers in NLP have recently begun
to develop lexicons of connotations
(\cite{feng-bose-choi:2011:EMNLP,fengacl2013}), i.e., words associated
with positive and negative polarities, out of context.  For example,
{\it war} has negative connotation and {\it sunshine} has positive
connotation.
Conceptually, though, {\it war} (used with an objective sense
referring to physical warfare) is not itself subjective.  While
it's likely that \textit{war} distributes more frequently with negative subjective expressions,
positive subjectivity
is certainly possible, as in {\it Ghenghis Kan likes war}.
When we
refer to {\it subjectivity} or {\it subjective expressions}, we
mean that, pragmatically, attitudes {\bf are} attributed to sources in
that context as part of the message being conveyed.

\subsubsection{Inferred Private States and Opinion Implicatures}
Consider the following example from the MPQA corpus (\cite{mpqa}):

\small
\begin{quote}
I think people are {\bf happy} because Chavez has fallen.
\end{quote}
\normalsize

\textit{Happy} clearly indicates (according to the writer) a positive
sentiment of the people toward Chavez's falling.  At the same time, we
might also \textit{infer} a negative sentiment of the people toward Chavez
himself, since they
are happy about an event harmful to him.

We address private states inferred from other private states, where
the attitude type of both is sentiment.
Inference is initiated by explicit sentiment subjectivity, either toward a gfbf
event (as in this example), or toward the agent or object of a gfbf
event (examples are given in Sections \ref{objectgfbf},
\ref{sourceattitude}, and \ref{agentgfbf}).

We borrow the term {\it implicature} from linguistics, specifically
{\it generalized conversational implicature}.
\cite{grice1967,grice1989} introduced the notion
to account for how more can be pragmatically communicated than what is
strictly said -- \\
\underline{what is implicated} vs. \underline{what is said}
(\cite{implicaturelanguage2012}).  Generalized conversational
implicatures are cancellable, or defeasible.

Analogously, we can treat {\bf subjectivity} as {\bf part of \underline{what is
said}},\footnote{While the focus in the literature on \underline{what is said}
is semantics, Grice and people later working on the topic acknowledge
that \underline{what is said} must include pragmatics such as co-reference
and indexical resolution (\cite{implicaturelanguage2012}), and
subjectivity arises from deixis
(\cite{bruderwiebe1995,steinwright1995}).  However, as long as \underline{what is
said} is conceived of as only truth evaluable propositions,
then it is not exactly the notion for our setting.
}
and the {\bf private-state inferences}
we address to be {\bf part of \underline{what is implicated}}.
Opinion implicatures are default inferences
that may not go through in context.

Though much previous work on sentiment analysis is relevant to
our work, there is almost no previous work in NLP that focuses on opinion implicature.
The closest is research on plot units and affect interpretation
(\cite{lehnert:plotlunits:cogsci:1981,daume10plotunits-emnlp}) and
related work on inferring goals in the interpretation of narratives
(\cite{schank1977,wilensky:pam}).
Recent research in linguistics, however, investigates one of
the opinion implicature rules that we propose (see RS1
in \ref{overview:inference}). In particular, \cite{reschkeAnand2011} carry
out a corpus study of the application of the inference rule via
sentences that match a set of fixed linguistic templates and find
that, in general, the predicted inferences hold in context.
Their results bode well for our general approach.

\subsubsection{Benefactive/Malefactive Events and States}
This work addresses sentiments toward, in general,
states and events which positively or negatively
affect entities.  Various lexical items and semantic roles
evoke such situations.
We focus on one clear case
that occurs frequently in opinion sentences:
$\langle$\textit{agent, event, object}$\rangle$
triples, where $event$ negatively or positively
affects the $object$.  Our terms for such events are {\it benefactive}
and {\it malefactive}, or, for ease of writing, \textit{goodFor} and \textit{
badFor} (hereafter \textit{gfbf}).
Focusing on this clear case
will make developing a fully automatic, end-to-end system
more feasible.   As other cases become clear, they may be
incorporated into the framework in the future.

A {\sc goodFor} event is an event that positively affects an entity
(similarly, for {\sc badFor} events).  Note that gfbf objects
are not equivalent to benefactive/malefactive semantic roles.  For
example, \textit{She baked a cake for me}: \textit{a cake}
is the object of {\sc goodFor} event \textit{baked} (creating
something is good for it (\cite{anandreschke2010}), while \textit{me}
is the filler of its benefactive semantic role
(\cite{zuniga2010benefactives}).

A {\it reverser} is an expression that
that reverses the polarity of a gfbf event (from {\sc badFor} to {\sc
goodFor}, or vice versa).

We have annotated a corpus with gfbf information and the
speaker's sentiments toward the agents and objects of gfbf events
(\cite{dengacl2013}).\footnote{Available at http://mpqa.cs.pitt.edu}

\subsection{Opinion Inference}
\label{overview:inference}

In this section, we introduce opinion inferences by stepping through
examples of inference.
The system includes default inference rules which apply if
there is no evidence to the contrary.  The system requires as input
explicit sentiment and gfbf information (plus
any evidence that is contrary to the inferences).\\

\noindent
Consider this example:\\

\noindent
{\bf {\sc Ex(1)}} Why would [President Obama]
  \textbf{support} [health care reform]? Because [reform] could \textbf{lower} [\emph{skyrocketing} health care costs], and \textbf{prohibit} [private insurance companies] from \textbf{overcharging} [patients].\\

\noindent
Suppose an explicit-sentiment analysis system recognizes only one
explicit sentiment expression,
(\textit{skyrocketing}). There are several gfbf events - {\it lower},
{\it overcharge}, and {\it support}. \\

\noindent
The input to the system for {\sc Ex(1)} is the
following. The first line, for example, represents a gfbf event,
$E_1$, whose agent is {\it reform} and whose object is {\it costs}.  The gfbf term
is {\it lower}, and the polarity of the gfbf event is {\sc badFor}.  The
last line represents the fact that the writer's sentiment toward the
costs is negative.\\

\noindent
$E_1$: $\langle$reform, lower:{\sc badFor}, costs$\rangle$\\
$E_2$: $\langle$reform, prohibit:{\sc reverse}, $E_3$ $\rangle$\\
$E_3$: $\langle$companies, overcharge:{\sc badFor}, patients$\rangle$\\
$E_4$: $\langle$Obama, support:{\sc goodFor}, reform$\rangle$\\
$S_1$: sent(writer,costs) = neg\\

There are 10 rule schemas implemented in the rule-based system.  Since
this is an overview, we give four somewhat simplified schemas here.

In the following, \textit{sent(X,$\alpha$) = $\beta$}
means that X's sentiment toward \textit{$\alpha$} is
\textit{$\beta$}, where \textit{$\alpha$} is a {\sc goodFor} event, a
{\sc badFor} event, or the agent or object of a gfbf event, and
$\beta$ is either $positive$ or $negative$ ($pos$ or $neg$, for short).
P $\rightarrow$ Q means to infer Q from
P.\\

\noindent
{\bf RS1: sent($S$,gfbf event) $\rightarrow$ sent($S$,object)}\\
1.1 sent($S$,{\sc goodFor}) = pos $\rightarrow$ sent($S$,object) = pos\\
1.2 sent($S$,{\sc goodFor}) = neg $\rightarrow$ sent($S$,object) = neg\\
1.3 sent($S$,{\sc  badFor}) = pos $\rightarrow$ sent($S$,object) = neg\\
1.4 sent($S$,{\sc  badFor}) = neg $\rightarrow$ sent($S$,object) = pos\\

\noindent
{\bf RS2: sent($S$,object) $\rightarrow$ sent($S$,gfbf event)}\\
2.1 sent($S$,object) = pos $\rightarrow$ sent($S$,{\sc goodFor}) = pos\\
2.2 sent($S$,object) = neg $\rightarrow$ sent($S$,{\sc goodFor}) = neg\\
2.3 sent($S$,object) = pos $\rightarrow$ sent($S$,{\sc badFor}) = neg\\
2.4 sent($S$,object) = neg $\rightarrow$ sent($S$,{\sc badFor}) = pos\\

\noindent
{\bf RS3: sent($S$,gfbf event) $\rightarrow$ sent($S$,agent)}\\
3.1 sent($S$,{\sc goodFor}) = pos $\rightarrow$ sent($S$,agent) = pos\\
3.2 sent($S$,{\sc goodFor}) = neg $\rightarrow$ sent($S$,agent) = neg\\
3.3 sent($S$,{\sc  badFor}) = pos $\rightarrow$ sent($S$,agent) = pos\\
3.4 sent($S$,{\sc  badFor}) = neg $\rightarrow$ sent($S$,agent) = neg\\

\noindent
{\bf RS4: sent($S$,agent) $\rightarrow$ sent($S$,gfbf event)}\\
4.1 sent($S$,agent) = pos $\rightarrow$ sent($S$,{\sc goodFor}) = pos\\
4.2 sent($S$,agent) = neg $\rightarrow$ sent($S$,{\sc goodFor}) = neg\\
4.3 sent($S$,agent) = pos $\rightarrow$ sent($S$,{\sc badFor}) = pos\\
4.4 sent($S$,agent) = neg $\rightarrow$ sent($S$,{\sc badFor}) = neg\\

\noindent
Applying the rule schemas to {\sc Ex(1)}, we have the following inferences.\\

\noindent
In $E_1$,
from the negative sentiment expressed by \textit{skyrocketing} (the writer is negative toward the costs because they are too high),  and
the fact that {\it costs} is the object of a {\sc badFor}
event ({\it lower}), Rule2.4 infers \textbf{a positive attitude toward $E_1$} .

Now, Rule3.3 applies.  We infer the writer is \textbf{positive toward
  the \textit{reform}}, since it is the agent of $E_1$, toward which
the writer is positive.

$E_2$ is an event that reverses the polarity of its object,
$E_3$. $E_3$ is the event of companies overcharging patients.
And, while companies overcharging patients is
{\sc badFor} them, reform {\bf preventing} companies from doing so is
{\sc goodFor} patients.  Thus, compositionally, we have a new event: \\

\noindent
$E_{3R}$: $\langle$reform, {\sc goodFor}, patients$\rangle$\\

Above, we inferred that the writer is positive toward \textit{reform},
the agent of $E_{3R}$. By Rule 4.1, the writer is \textbf{positive toward
$E_{3R}$}.  Since the writer is positive toward $E_{3R}$, and $E_{3R}$
is {\sc goodFor} patients, Rule 1.1 infers that the writer is \textbf{positive toward patients}.

Turning to $E_4$, \textit{support health care reform} is {\sc
goodFor} reform.  We already inferred the writer is positive
toward {\it reform}.  Rule 2.1 infers that the writer is {\bf positive toward
$E_4$}. Rule 3.1 then infers that the writer is {\bf positive toward the agent of $E_4$, \textit{Obama}}.

In summary, RS1-RS4 infer that the writer is positive toward $E_1$, health care
reform, $E_{3R}$, patients, $E_4$, and Obama.
Thus, from a single explicit {\bf negative} sentiment, the system infers several
{\bf positive} sentiments.\\

We now apply RS1-RS4 to an example from
a political blog on the site redstate.com.\\

\noindent
{\sc Ex(2)}
It is no surprise then that [MoveOn] would \textbf{attack} [Senator McCain].\\

\noindent
In the context of the blog containing it, this sentences expresses
negative subjectivity toward the event.
(The linguistic clues, which are ambiguous, are fronting,
\textit{surprise}, and \textit{then}
(\cite{wiebe1994}).  We return to this example in Section
\ref{exploring}.) \\

\noindent
Following are the inputs for {\sc Ex (2)}.\footnote{Note that the word
sense of \textit{attack}
in {\sc EX(2)} is
subjective (\cite{wiebe-mihalcea:2006:COLACL}), so there is
also negative subjectivity of MoveOn toward McCain in the sentence, nested in the
subjectivity of the writer.  To keep this example from being too
complex, we did not include it in the input to the system.
When we add it to the input and re-run the
system, the new inferences are consistent with the previous ones, and
the additional sentiment reinforces the overall interpretation.} \\

\noindent
$E_5$: $\langle$MoveOn,attack:{\sc badFor},McCain$\rangle$\\
$S_2$: sent(writer,$E_5$) = neg\\

Among RS1-RS4, two rules apply: Rule 1.4 infers that the writer is positive toward
McCain,
and Rule 3.4 infers that the writer is negative toward MoveOn. \\

So far in this section, the source of all the private states
(both explicit and inferred)
has been
the writer.
We now consider private
states whose sources are entities mentioned in the sentence.

As will be seen below, in addition to the inferences laid out above,
the system also {\bf ascribes attitudes} to {\bf the agents} of the
gfbf events.  For Ex(2), through a sequence of inferences, the system
infers that the writer believes that (1) MoveOn is negative toward
McCain, (2) MoveOn intentionally performed the action that is {\sc
badFor} McCain, and (3) MoveOn wanted to (i.e., is positive toward)
perform the action.  This is accomplished by the fuller set of rules
used by the system (as mentioned above, RS1-RS4 are simplifications).
Importantly, as will be seen in Section \ref{blocking}, the rules used
to infer that the writer believes (1)-(3) are the {\bf same
rules} involved when inferences toward agents are {\bf blocked}. The
inferences are blocked by evidence that breaks the inference chains
leading to (1)-(3).  Thus, we do not need separate rules for {\bf richer
inference}, on the one hand, and for {\bf defeating implicatures}, on the
other; defeated inference occurs when expected inference is blocked
(\cite{levinson1983pragmatics,schank1977}).

Further, if a rule matches a sentiment or event that is the target of
a private state, the nesting structure is preserved when generating
the conclusions.  We say that inference is carried out in {\it private
state spaces}.  In addition to drawing {\bf conclusions} within private
state spaces, the system {\bf defeasibly infers that the rule premises and
assumptions are in those spaces as well}.  private-state spaces are
described in Section \ref{psspaces}.

Note that rules are applied repeatedly until no new conclusions can be
drawn.  Thus, even though there are a total of only ten rule schemas,
the number of inferences may be quite high.\\

\noindent
Consider Ex(3), in which the writer expresses a sentiment toward a sentiment.\\

\noindent
{\bf {\sc Ex}(3)}
However, it appears as if the international community (IC)
is tolerating the [Israeli] \textbf{campaign of suppression against}
[the Palestinians]. \\

\noindent
The input annotations are the following: \\

\noindent
$E_6$: $\langle$Israeli,suppression:{\sc badFor}, palestinians$\rangle$\\
$S_3$:  sent(IC,$E_6$) = positive\\
$S_4$:  sent(writer,$S_3$) = negative\\

\noindent
The IC is positive toward the event in the sense that they tolerate
it. \textit{However} and \textit{appears as if}  are clues that the writer is
negative toward IC's positive sentiment. \\

\noindent
The following are the
sentiments inferred {\bf just toward the entities} in the sentence; note that
many of the sentiments are nested in private-state spaces: \\ \\

{\it  writer is positive toward the Palestinians}

{\it  writer is negative toward Israel}

{\it  writer is negative toward international community}

{\it writer believes that Israel is negative toward the Palestinians}

{\it writer believes that international community is negative toward the Palestinians}

{\it writer believes that international community is positive toward Israel}

{\it writer believes that international community believes that Israel
  is negative toward the Palestinians}  \\

This section presented an overview.
Four rule schemas were presented, which show that inference may
proceed from sentiment toward events to sentiment toward
entities (RS1 and RS3), or from sentiment toward entities
to sentiment toward events (RS2 and RS4).  We introduced the idea that
richer inference provides the basis for defeasibility, and the notion
that inference is carried out in private-state spaces.
In total, the
rule-based system has 10 rule schemas; all of the inferences are made
via (repeated) applications of rules in these schemas.  If a rule matches a
sentiment or event that is the target of a private state, the nesting
structure is preserved.
All of the inferences carried out by the system are default rules
which may be blocked by evidence to the contrary.

\section{Data Structure}
\label{datastructure}
The system builds a graphical representation of
what it knows and infers about the meaning of a sentence.

Sentences are processed independently;
so, the graphs for two different sentences do not share any nodes. 

The system begins by building a graphical representation of the inputs for a
sentence.  Below is example input for a sentence.  
Note that the the id's in the input are only used
by the system to build the initial graph.  The nodes of the graph are numbered
independently from the input id's.  

\small
\textsf{ \\ 
\noindent
\noindent
"Mayor Blooming idiot urges Congress to vote for gun control."   \\
\noindent
E1 gfbf $\langle$Congress, goodFor (voting for), gun control$\rangle$ \\
S1 subjectivity $\langle$Mayor-Blooming-idiot, positive sentiment (urges), E1$\rangle$ \\
B1 privateState $\langle$writer, positive believesTrue (""), S1$\rangle$ \\
S2 subjectivity $\langle$writer, negative sentiment (blooming idiot), Mayor-Blooming-idiot$\rangle$  \\
}  \normalsize 

The system builds these nodes (as printed by the system):

\small
\textsf{ \\ 
\noindent
\noindent
\noindent 43 writer positive believesTrue \\
\noindent   \indent  41 Mayor-Blooming-idiot positive sentiment  \\
\indent \indent        38 Congress voting for gun control \\
\noindent 45 writer negative sentiment \\
\indent    42 Mayor-Blooming-idiot \\
}  \normalsize 

The system's printout does not show all the structure of a node.
Consider node 41.  It has a {\it source} edge to the node representing
Mayor-Blooming-idiot, and a {\it target} edge to node 38, which in turn has
an {\it agent} edge to the node representing Congress and a {\it
  goodFor} edge to the node representing gun control (for convenience,
there is also an edge labeled {\it object} from node 38 to the node
representing gun control).  The nodes also have attributes which
record, e.g., what type of node it is (node 41 is a privateState and node
38 is a gfbf), polarity (if relevant), etc.

The graph is directed.  For example, node 38 is a child of 41, which is
a child of 43.

A specification for the input is that each root node must be a sentiment or believesTrue node
whose source is the writer. 

Inference proceeds by matching rules to the graph built so far and,
when a rule successfully fires, adding nodes to the graph.

The following sections give the knowledge representation scheme for
nodes (in \ref{kr}), specify the form and meaning of explicit inferences rules (in
\ref{rulespec}), and describe a mechanism for attitude ascription (in
\ref{psspaces}).  The various specifications
give us a graphical data structure consisting of chains of
believesTrue and sentiment nodes, where nodes are linked in a chain
via the {\it target} relationship.
The source of the root of
each chain is the writer, and the target of the rightmost node of a
chain is not a believesTrue or sentiment node.  
As we will see in \ref{psspaces}, a node that is the target of the
rightmost node on a chain is considered to be in the {\it private-state space}
defined by that chain.

\section{Knowledge Representation Scheme}
\label{kr}

Node X having attribute A with value V means A(X,V). \\

Node P having child C, where the label of the edge from P to C is L,
means L(P,C).\\

A node represents a concept of something.  Each node has a {\it type}
attribute, specifying what type of thing the node represents.\\

Below, we present each type of node, its attributes, and children.
\begin{itemize}
\item {\it Type=anim:}
An animate agent.  A node with no attributes or children. \\

\item {\it Type=thing:} A thing.  Like anim, a node no attributes or children.\\

\item {\it Type=state:}  a state other than a privateState. 

{\bf Children:} experiencer, object (type: anim or  thing).

{\bf Attributes:} None.\\

\item {\it Type=event:}  an event other than a gfbf. 

{\bf Children:} agent, object (type: anim or  thing).

{\bf Attributes:} None.\\

\item {\it Type=gfbf:} An event with benefactive or malefactive effect on
something or someone. 

{\bf Children:} agent, object (type: anim or  thing).  If the event is
goodFor (badFor) the object, then the node also has a goodFor (badFor)
edge to the
object.  

{\bf Attributes:} None.\\

\item {\it Type=ideaOf:} The idea of a gfbf.

{\bf Child:} ideaObject (type: gfbf).

{\bf Attributes:} None. \\

X ideaObject Y means that X is the idea of Y.\\

\item {\it Type=p(x):} 

{\bf Attribute:} property (isBad, isGood, isTrue, isFalse, should,
or shouldNot).

{\bf Child:} x (type: anim, thing, ideaOf, agreement, privateState)\\

\item {\it Type=agreement:} 

{\bf Attribute:} polarity (positive or negative).

{\bf Children:} source (type: anim), withWhom (type: anim), target
(type: p(x)). \\

Suppose X is an agreement node with polarity = negative; source S;
withWhom W; and target p(x).  This means that S disagrees with
W that p(x).   The disagreement is from S's perspective
(i.e., according to S, W disagrees with him about p(x)).\\

\item {\it Type=privateState:}

{\bf Attributes:} attType (believesTrue, sentiment, intends, or believesShould);
polarity (positive or negative).

{\bf Children:} source (anim), target (which types are allowed depends
on the attType of the node). \\

The source is the immediate source of the private state.
Nested sources are not represented explicitly; they are defined
dynamically by the
nesting of privateState and agreement nodes created so far for the
current sentence. \\

The attitude types are the following:

\begin{itemize}
\item {\it attType=believesTrue:}

{\bf Target types}: privateState, agreement, p(x), or gfbf. \\

What S positive believesTrue T means depends on the type of T.

Note that privateStates, agreements, and p(x)'s are all propositions.
It makes sense to say, for example, that the writer believes that X is positive toward
something (a private state), the writer believes that X disagrees with
Y about something (an agreement), or that the writer believes that
something is
good (a p(x)). Thus, if T is one of those (i.e., not a gfbf), then
S positive believesTrue T simply means that S believes that T. \\

Events, on the other hand, are not themselves propositions.
For an event to be
the object of {\it believes that}, it needs to be coerced into a
proposition.   We handle this by saying that the source has to believe
{\it something} about the event. 
Thus, if T is a gfbf, then S positive believesTrue T means that
there exists some p such that S
believes that p(T).  Typically, we don't commit to what ``p'' is.  The
exception is p = ``substantial''. S positive believesTrue
substantial(T) means that S believes that T is ``real,''
i.e., that the
event happened or will happen.  That's what we {\it mean} by the sentence ``John believes that
Mary killed Bill,'' i.e., John believes that the event
happened.\footnote{In general, in developing the framework, I tried to
  avoid requiring judgments from the future NLP system as to
whether events are substantial or not.  Thus, p=substantial is only
provided on the input when there is a reason to.  As you will see
below, sentiment inferences toward targets and animate agents do not hinge on
whether the gfbfs are substantial.  Further, by allowing p to be
unspecified, a belief toward a gfbf in the input does not
commit to the belief that the event is substantial/real.}\\

Finally, let's consider what S {\bf negative} believesTrue T means.
It means (A) or (B):
\begin{itemize}
\item[(A)]
S does not believe that T, in the sense that T is not in S's belief space.
Among the things that S believes are true, T is not one of them.
S may not believe anything about T.
\item[(B)]
S does not believe that T, in the sense that S believes that not T, i.e., S believes that T
is false.
\end{itemize}
Section \ref{negBelMeaning} addresses this issue in some detail (that
section may be safely skipped for those not interested).
Suffice it to say here that we have one representation
that means (A) or (B); our knowledge representation does not
distinguish between the two cases.  We could extend our KR to
distinguish between them if needed in the future.\\

\item {\it attType=sentiment:} 

{\bf Target types:} gfbf, p(x), privateState, anim, thing,
agreement, or
ideaOf.\\

S positive sentiment toward T means that S is positive toward T
(similarly for negative sentiment).\\

\item {\it attType=intends}

{\bf Target type:} gfbf.   \\

S positive intends T means that S intends or intended to do T.\\

S negative intends T means that S does not/did not intend to do T.  It
does not mean that S intends to do the opposite of T.  That is,
``negative intends'' is the absence of intention.
\\

\item {\it attType=believesShould:}

{\bf Target type:} gfbf.   \\

S positive believesShould T means that S believes that the agent of T should do T.\\

S negative believesShould T means that S believes that the agent of T should not do T.\\
\end{itemize}
\end{itemize}
\section{Inference}
\label{inferencesGeneralDescription}
This section describes the inferences carried out by the system.
The basis of the inference is, as expected, a set of rules (described
next in \ref{rulespec}).  But, the rules alone are not sufficient,
because the inferences are carried out
in the context of private states.  Once a rule fires, what attitudes toward
the conclusions should be ascribed, and to whom?
These questions are addressed in \ref{psspaces}.
Finally, all of the reasoning is done by default.   The final
subsection, \ref{blocking}, describes the types of negative evidence which may block
an inference.

It should be emphasized that sentences are processed independently
from one another; private states are not carried over from sentence
to sentence.  However, one could process an entire discourse by treating
it as one long sentence.

Further, all of the reasoning mechanisms and rules assume that, within a
private-state space, all the attitudes of a particular source are
consistent with each other.  For example, within a private-state
space, there cannot be two nodes with the same sources, targets, and
attitude types, but different polarities (we return to this point
in Section \ref{blocking}).

\subsection{Rules}
\label{rulespec}

The rules are default rules:  a conclusion cannot be drawn if there is
evidence against it. 

A rule may include assumptions.
For example,
suppose a rule would successfully fire if S believes P for some S
mentioned in the sentence.  If the writer believes P, and there is no
evidence to the contrary, then we'll assume that S believes it as
well, if a rule ``asks us to''.

Thus, our rules are conceptually of the form: \\

P1,...,Pj: A1,..,Ak/Q1,...,Qm \\

where the Ps are preconditions, the As are assumptions, and the Qs are
conclusions.   For the Qs to be concluded,  the Ps must already hold; there
must be a basis for assuming each A; and there must be no evidence
against any of the As or Qs.

Here is an example rule (as displayed by the system), which has one
precondition and one conclusion.

\textsf{ \\ 
\noindent
\indent S sentiment toward A goodFor/badFor T $\Longrightarrow$   \\
\indent S sentiment toward the idea of A goodFor/badFor T   \\
} 

Assumptions are indicated using the term ``Assume'', as in this rule: 

\textsf{ \\ 
\noindent
\noindent
S sentiment toward A goodFor/badFor T, where A is a thing \&   \\
\noindent
(Assume S positive believesTrue substantial) A goodFor/badFor T $\Longrightarrow$   \\
\noindent
S sentiment toward A   \\
} \\ 
The first line contains a precondition, the second an assumption,
and the third a conclusion.
\subsection{Private-State Spaces}
\label{psspaces}
Inference is carried out in \textit{private-state spaces}.
The system's knowledge is represented as 
chains of
believesTrue and sentiment nodes, where nodes are linked in a chain
via the {\it target} relationship.
The source of the root of
each chain is the writer, and the target of the rightmost node of a
chain is not a believesTrue or sentiment node.  
A node that is the target of the
rightmost node on a chain is considered to be in the {\it private-state space}
defined by that chain.

Other than private states of the writer, all propositions and events must be the target of {\it some} private state.  In the
simplest case, the proposition or event is simply positive believedTrue by the
writer.

We want to carry out inferences within private-state spaces so that, for
example, from S positive believesTrue P, and P $\rightarrow$ Q, the
system may infer S positive believesTrue Q.  However, we are working
with sentiment, not only belief, and we want to
allow, as appropriate, these types of inferences: from S sentiment
toward P, and P $\rightarrow$ Q, infer S sentiment toward Q.  For example, If
I'm upset my computer is infected with a virus, then I'm also upset
with the consequences (e.g., that my files are corrupted).

Not all inferences are carried out via explicit rules; those such as
the one just described involve ascription in private-state spaces.

The Ps, As, and Qs of the rules match
nodes in the graph where
their ancestors are all sentiment and believesTrue nodes.  A
private-state space is defined by a path where the root is a
believesTrue or sentiment node whose source is the writer, and each
node on the path is a believesTrue or sentiment node.  Two paths define the
same private-state space if, at each corresponding position, they have the same attType,
polarity, and source.  P is {\it in} a private-state space if P is the
target of the rightmost node on a path defining that space.

So, suppose that (1) the preconditions, Ps, of a rule hold, (2) there is a basis
to make each assumption A, and (3) there is no evidence against any of
the As or the Qs.
For each private-state space containing all of the Ps, nodes are built to place
the As and Qs in that space as well.  We call this process {\it
space extension}.

However, if there is a negative believesTrue on the
path defining a space, space extension is not carried out.  I.e., this
is one way an inference may be blocked.  The other two ways inferences
may be blocked are described in the following
subsection.
\subsection{Blocking of Inferences}
\label{blocking}
There are two additional ways inferences may be blocked:
within the context of a private-state space, and based on evidence
from outside the rule-based system.

\subsubsection{Explicit Private States in Private-State Spaces}

An inference is blocked if space expansion fails, i.e., if there is
not a single private-state space 
(1) whose path does not include any
negative believesTrue nodes, (2)
which includes all of the Ps, and
(3)
to which it would be valid to add all of the As and Qs.
The third item still needs to be explained.  It would not be valid to
add a proposition P to a private-state space S if:
\begin{enumerate}
\item
The rightmost node of S is a negative believesTrue with target P.
(This
is redundant with the restriction that the private-state space path cannot
contain negative believesTrue nodes; we leave this as a separate
condition for now.)
\item
There is a node P1 already in S such that the sources and attitude types of P
and P1 are the same; their targets have the same structure;
but their polarities are opposite.  For example, in the same space, we
cannot have both a positive and a negative sentiment toward the same
target.\footnote{If in the future it appears there is a need to do so,
that would mean that the targets need to be more fine-grained;
i.e., if we want a space to include both a positive and negative
sentiment toward something, we need to split into two sentiments,
one toward the positive aspect and one toward the negative aspect.}
\end{enumerate}

\subsubsection{Evidence Against}
In a practical setting, where the input will come from an NLP system
and not manual annotations, it will generally not be reasonable to
expect a system to make fine-grained private-state space distinctions.

Thus, we also have ``evidence'' nodes on the input, which indicate
attitude type, polarity, and target, but which are not placed in
private-state spaces.   Rule inferences are blocked if
there is evidence against an assumption or conclusion.  That is, if
there is evidence from the NLP system against, e.g., Q1, then the
assumption or inference of Q1 is blocked from going through in any of the private-state spaces.

Importantly, the mechanisms for blocking inferences based on explicit
private states in private-state spaces {\bf cannot} be replaced by 
inference blocking based on evidence recognized by an NLP system.  The
private-state-space-blocking mechanism is needed for the internal reasoning
of the system, i.e., for the proper coordination of inference rules and
space extensions.

\subsection{Attitude Ascription}
\label{attitudeAscription}
This section provides more detail about attitude ascription.

There are two places where attitude ascription is performed:
when making assumptions of the As of a rule,
and during space extension.

First, consider assumptions.
There are two bases for making an assumption.  First, we have a basis for
assuming S positive
believesTrue P if the writer positive believesTrue P.  Second, we
have a basis for assuming S has an attitude of type AT toward
target T if S already has an attitude toward T which is {\it not} of
type AT (of course, as long as the existing attitude is not a
negative believesTrue).  For example, we have a basis for assuming
S positive believesTrue T if S has a sentiment toward T.

Remember, assumptions may later be blocked; they are actually
concluded (in appropriate private-state spaces) only if the entire rule
successfully fires.

Turning to space extension, there is one more piece of information to
give. 

As described above, if a rule successfully fires, the As and Qs are
placed in all of the private-state spaces that already contain all the
Ps.

There is one more step:  for each of those spaces which includes
a sentiment to the left of the path's end, the Ps, As, and Qs are added to the
spaces created by replacing sentiments with beliefs on the path.
The idea is, if S has a sentiment toward T, then S must have a
positive believesTrue attitude toward T.  Of course, people may have
sentiments toward events that never happen.  But that's fine.
Remember, if T is a gfbf,
S positive believesTrue T means that S positive believesTrue p(T), for
some p.  I.e., S believes {\it something} about the gfbf; this
does not imply that S believes the event really happened (i.e., we are
not committing to p == substantial).  

The reader may notice when going through
the examples in Section \ref{examples} that there are some additional beliefs the
system may be warranted in inferring.  For the interested reader, these are discussed in
Section \ref{extensionsThatCouldBeAdded}.  However, that section may
be safely skipped for those not interested.

\section{Compositionality}
\label{compositionality}

The main type of compositionality performed by the system is
composition of {\it influencers} and gfbfs.  Influencers are either
{\it reversers} or {\it retainers}.
Here are some examples.
In \textit{John didn't kill Bill}, the negation is a reverser influencer on the
gfbf event of killing Bill.  The system creates a new gfbf event, \textit{John goodFor
Bill.}  The new gfbf is the one that participates in the
inferencing; the original one from the input is ignored by the
system from this point forward.

In \textit{John helped Mary to save Bill}, \textit{helped} is a retainer
influencer.  The system builds a new gfbf:
\textit{John goodFor Bill}.

Now, because influencers are reversers or retainers, they are also,
conceptually, gfbf's!  Further, there may be chains
of them:  for example, the structure of \textit{He stopped trying to kill Bill}
is a reverser with object retainer (\textit{trying}) which has object \textit{badFor Bill}.  In our
gfbf and speaker-attitude annotation scheme (\cite{dengacl2013}), we opted to treat
all such chains of length {\it N} as {\it N-1} influencers followed by
a single gfbf.  

The system then starts from the gfbf, and follows the
backward chain of influencers, creating new gfbf nodes as it goes.
Only the final one is added to the set of nodes being retained to
represent the sentence.

The question arises, how should evidence nodes be handled in such
cases?  It turns out that creating appropriate new evidence nodes as the new
gfbf nodes are created during compositionality is straightforward.

For example:

\small
\textsf{ \\ 
\noindent
\noindent
"""Oh no! The tech staff tried to stop the virus, but they failed.   \\
\noindent
\noindent 115 writer negative sentiment \\
\indent 114 the tech staff $\langle$reverse$\rangle$  \\
\indent \indent 111 the tech staff stop the virus \\
\noindent 118 There is evidence that the following is not substantial \\
\indent 111 the tech staff stop the virus \\
\noindent 117 There is evidence that the following is intentional: \\
\indent 111 the tech staff stop the virus \\
}  \normalsize

In the input, we have a gfbf that is intentional (they tried
to stop the virus), but is not substantial (they failed to do so).\\

The new gfbf:

\small
\textsf{ \\ 
\noindent
\noindent
\noindent
\textbf{influencer} node:  \\
\noindent  \textit{115 writer negative sentiment}   \\
\indent   114 the tech staff $\langle$reverse$\rangle$  \\
\indent \indent   111 the tech staff stop the virus \\
\noindent
\textbf{new gfbf} node:  \\
\noindent  \textit{1262 writer negative sentiment}  \\
\indent   1259 the tech staff goodFor the virus  \\
\noindent
\textbf{new evidence} node:   \\
\noindent
\noindent 1260 There is evidence that the following is not intentional:  \\
\indent 1259 the tech staff goodFor the virus  \\
\noindent 1261 There is evidence that the following is substantial \\
\indent 1259 the tech staff goodFor the virus \\
}  \normalsize

The new gfbf is 1259, the tech staff goodFor the virus.  The new
evidence nodes are 1260 and 1261; they have opposite polarities from
their counterparts in the input.
Gfbf event 1259 is substantial, because a good thing did happen to the virus:  it wasn't
stopped.  But it is not intentional:  the tech staff didn't try to do something
goodFor the virus; in fact, they tried to do something badFor it (stop
it).\\

The next point is not concerned with compositionality, per se, but this is a good
place to point out another case of goodFor/badFor
relationships the system can handle if appropriate entries are made
in the lexicon:

\small
\textsf{ \\ 
\noindent
\noindent
``The people are angry that the leader deprived the children of food''   \\
\noindent
194 writer positive believesTrue   \\
\indent 192 the people negative sentiment   \\
\indent \indent 193 the leader deprived the children of   \\
\indent \indent \indent \indent 191 food; which is goodFor the children  \\ 
}  \normalsize 

The lexical entry for \textit{deprive} notes that depriving is
badFor the object (here, the children).  \textit{Deprive} also takes another
semantic role, here filled by \textit{food}; \textit{deprive}'s entry states that the
filler of that second role is goodFor the filler of the object role,
whatever it is.  If someone is deprived of something, then that thing
is good for them (\cite{Moilanen2007}, among others, have rules for
such cases).

%
%
%
%

\section{The Rules}
\label{therules}

This section gives the rules.
They are described as we step through examples below; here, we address two
questions.

First, why {\bf ideaOf} A goodFor/badFor T?  Because the purview of this
work is making inferences about attitudes, not about events
themselves.  Conceptually, {\it ideaOf} is important for making the
knowledge representation
scheme consistent.  An intuition is that it coerces an event into
an idea, raising it into the realm of private-state spaces.

Second, why are assumptions sometimes (but not always) included in
rules?  The explanation is that, for a rule to fire, all of the
preconditions must be at the ``same level'' with respect to
private-state spaces, in order for the operation of adding the conclusions to
private-state spaces to be well defined. 
Consider a rule: P1 \& P2 $\rightarrow$ Q.  As discussed in Section
\ref{attitudeAscription}, Q is added to each private-state space that
already contains both P1 and P2.  If P1 is an attitude and P2 is not,
then \textit{adding Q to the spaces that contain both P1 and P2} does
not make sense.  In this case, the rule includes an assumption toward
P2 that nests P2 in a private-state space, leveling things out.

The ten rule schemas are the following:

\small
\textsf{ \\ 
\noindent
\noindent
\fbox{rule8}   \\
\indent  S positive believesTrue A goodFor/badFor T \& 
S sentiment toward T   \\
 \indent $\Longrightarrow$ S sentiment toward A goodFor/badFor T \\ \\
\noindent
\fbox{rule1}   \\
\indent S sentiment toward A goodFor/badFor T \\
\indent $\Longrightarrow$ S sentiment toward the idea of A
goodFor/badFor T \\ \\
\noindent
\fbox{rule2}   \\
\indent S sentiment toward the idea of A goodFor/badFor T  \\
\indent $\Longrightarrow$  S sentiment toward T \\ \\
\noindent
\fbox{rule3.1}   \\
\indent
S1 sentiment toward S2 sentiment toward Z  \\
\indent  $\Longrightarrow$
S1 agrees/disagrees with S2 that isGood/isBad Z \&
S1 sentiment toward Z   \\ \\
\noindent
\fbox{rule3.2}   \\
\indent
S1 sentiment toward
  S2 pos/neg believesTrue substantial Z \\
\indent $\Longrightarrow$
S1 agrees/disagrees with S2 that isTrue/isFalse Z \&
S1 pos/neg believesTrue substantial Z   \\ \\
\noindent
\fbox{rule3.3}   \\
\indent
S1 sentiment toward
  S2 pos/neg believesShould Z   \\
\indent  $\Longrightarrow$
S1 agrees/disagrees with S2 that should/shouldNot Z   \&
S1 pos/neg believesShould Z   \\ \\
\noindent
\fbox{rule4}   \\
\indent
S1 agrees/disagrees with S2 that *  \\
\indent $\Longrightarrow$ 
S1 sentiment toward S2   \\ \\
\noindent
\fbox{rule6}   \\
\indent A goodFor/badFor T, where A is animate   \\
\indent $\Longrightarrow$  
A intended A goodFor/badFor T \\  \\
\noindent
\fbox{rule7}   \\
\indent
S intended S goodFor/badFor T \\
\indent  $\Longrightarrow$
S positive-sentiment toward ideaOf S goodFor/badFor T   \\ \\
\noindent
\fbox{rule9}   \\
\indent
S sentiment toward A goodFor/badFor T, where A is a thing \&   \\
\indent  \indent
(Assume S positive believesTrue substantial) A goodFor/badFor T   \\
\indent $\Longrightarrow$
S sentiment toward A   \\ \\
\noindent
\fbox{rule10}   \\
\indent
(Assume Writer positive believesTrue) A goodFor/badFor T \&   
T in connotation lexicon  \\
\indent  $\Longrightarrow$ 
Writer sentiment toward T   \\
} \normalsize

The following two rules are not general inference rules,
but rather capture the fact that, in English, explicit subjectivity is often expressed
toward an individual rather than a proposition or event in which the
individual participates.  Note that
they apply only to nodes that are directly from the
input.

\small
\textsf{ \\ 
\noindent
\fbox{rule5source}   \\
\indent
S1 sentiment toward S2 in the input \&   \\
\indent
(Assume S1 positive believesTrue) S2 privateState toward Z in input  \\
\indent $\Longrightarrow$ 
S1 sentiment toward  
  S2 privateState toward Z   \\ \\
\noindent
\fbox{rule5agent}   \\
\indent
S1 sentiment toward A in input \&   \\
\indent
(Assume S1 sentiment toward) A goodFor/badFor T in input \\
\indent  $\Longrightarrow$
S1 sentiment toward A goodFor/badFor T   \\
} \normalsize

\section{Control}
\label{control}
The rules are ordered. The process is
iterative.  On each iteration, all of the rules are applied, in the
same order.  Inference stops when none of the rules returns any nodes
during an iteration.

\section{Examples}
\label{examples}
We will present examples in a uniform way.  First, the input to the
system will be shown:

\small
\textsf{ \\ 
\noindent
\noindent
``Is it no surprise then that MoveOn would attack Senator McCain.!?''   \\
E1 gfbf         $\langle$MoveOn, badFor (attack,attack:lexEntry), Senator McCain$\rangle$ \\
S1 subjectivity $\langle$writer, negative sentiment (surprise \& then \& the question), E1$\rangle$ \\ 
}  \normalsize

\noindent
Then, the system's display of the graph built from the input:

\small
\textsf{ \\ 
\noindent
\noindent
4 writer negative sentiment   \\
\indent 1 MoveOn attack Senator McCain  \\ 
}  \normalsize
 
Finally, excerpts of the system's output will be shown, with
comments interspersed.  The full system output may be found in
Appendix A.\\

Below, we step through examples to motivate the rules and highlight 
significant reasoning chains.  To make this feasible, we 
created (or at least simplified) examples.
However, in the next subsection,
we show an example from the MPQA corpus, which illustrates
the case where annotators perceive
subjectivity which they cannot 
anchor to a clear text signal; the implicature rules provide an explanation.

\subsection{An example from MPQA corpus}
\label{mpqaexample}
Note that the types of inferences in this section are covered in turn
in the following subsections.

Consider the following sentence from MPQA:

\begin{quotation}
\noindent
Ex(5) [He] is therefore planning to \textbf{trigger} [wars] here and there to \textbf{revive} [the flagging arms industry].
\end{quotation}

There are two gfbf events in this sentence: $\langle$He, trigger,
wars$\rangle$ and $\langle$He, revive, arms industry$\rangle$. 

The manual inputs for this sentence are:

\small
\textsf{ \\
\noindent
\noindent
E1 gfbf         $\langle$He, goodFor (trigger), wars (war:lexEntry)$\rangle$\\
E2 gfbf         $\langle$He, goodFor (revive), flagging arms industry$\rangle$\\
B1 privateState $\langle$writer, positive believesTrue (""), E1$\rangle$\\
B2 privateState $\langle$writer, positive believesTrue (""), E2$\rangle$\\
}\normalsize

The input nodes built for the inference are:
 
\small
\textsf{ \\
\noindent
8 writer positive believesTrue \\
\indent 4 He revive flagging arms industry \\
\noindent 6 writer positive believesTrue \\
\indent 1 He trigger wars \\
}
\normalsize

The attribute \textit{lexEntry} in the input is a signal to retrieve
information from a lexicon.  In this case, the lexical entry for
\textit{war} indicates it has a negative connotation.  The first
inference is from connotation to sentiment:

\small
\textsf{ \\
\noindent
\textbf{\fbox{rule10}}   \\
\indent
(Assume Writer positive believesTrue) A goodFor/badFor T \&    \\
\indent
T's anchor is in connotation lexicon \\
\indent  $\Longrightarrow$ 
Writer sentiment toward T  
\vspace{-1em}   \begin{multicols}{2}
\noindent
\textbf{Assumptions:} \\
\noindent
6 writer positive believesTrue \\
\indent 1 He trigger wars \\
\vfill
\columnbreak
\noindent
\textbf{$\Longrightarrow$ Infer Node:} \\
\noindent
\noindent
17 writer negative sentiment \\
\indent 2 wars 
\end{multicols} \vspace{-1em}
}  \normalsize

From the writer's negative sentiment toward
\textit{wars}, the system infers a negative sentiment toward 
\textit{trigger wars}, since triggering wars is goodFor them:

\small
\textsf{ \\
\noindent
\textbf{\fbox{rule8}} \\
\indent  S positive believesTrue A goodFor/badFor T \& \\
\indent S sentiment toward T   \\
 \indent $\Longrightarrow$ S sentiment toward A goodFor/badFor T 
\vspace{-1em}   \begin{multicols}{2}
\noindent
\noindent
\textbf{Preconditions:} \\
\noindent
6 writer positive believesTrue  \\
\indent 1 He trigger wars \\
\noindent
17 writer negative sentiment \\
\indent 2 wars \\
\noindent
\vfill
\columnbreak
\noindent
\textbf{$\Longrightarrow$ Infer Node:} \\
\noindent
\noindent
28 writer negative sentiment \\
\indent 1 He trigger wars  
\end{multicols}  \vspace{-1em}
}
\normalsize

On the other hand, since the agent, \textit{He}, is animate and
there is no evidence to the contrary, the system infers that the
triggering event is intentional, and that \textit{He} is positive
toward the idea of his performing the event:

\small
\textsf{ \\
\noindent
\textbf{\fbox{rule6}}   \\
\indent A goodFor/badFor T, where A is animate   \\
\indent $\Longrightarrow$  
A intended A goodFor/badFor T   
\vspace{-1em}  \begin{multicols}{2}
\noindent
\textbf{Preconditions:} \\
\noindent
\textit{6 writer positive believesTrue}  \\
\indent  1 He trigger wars \\
\noindent
\textit{28 writer negative sentiment}  \\
\indent      1 He trigger wars  \\
\vfill  \columnbreak
\noindent
\textbf{$\Longrightarrow$  Infer Node:} \\
\noindent
\noindent
\textit{38 writer negative sentiment}   \\
\indent 20 He positive intends  \\
\indent \indent         1 He trigger wars \\
  \end{multicols} \vspace{-1em}
\noindent
\textbf{\fbox{rule7}}   \\
\indent  S intended S goodFor/badFor T \\
\indent  $\Longrightarrow$
S positive-sentiment toward ideaOf S goodFor/badFor T   \\
\vspace{-1em}  \begin{multicols}{2}  
\noindent
\textbf{Preconditions:}  \\
\noindent
\textit{21 writer positive believesTrue} \\
\indent     20 He positive intends  \\
\indent \indent       1 He trigger wars \\
\noindent
\textit{38 writer negative sentiment}  \\
\indent     20 He positive intends  \\
\indent \indent          1 He trigger wars  \\
\vfill  \columnbreak
\noindent
\textbf{$\Longrightarrow$  Infer Node:}   \\
\noindent
\noindent
\textit{41 writer negative sentiment}  \\
\indent   25 He positive sentiment \\
\indent \indent     26 ideaOf \\
\indent \indent \indent        1 He trigger wars \\
\end{multicols}   \vspace{-1em}
}
\normalsize

\noindent
Let us explain the output just shown for the application of rule 6.
The precondition match is node 1.  Node 1 appears in two
private-state spaces, \textit{writer positive believesTrue} (node 6) and
\textit{writer negative sentiment} (node 28).  Nodes 6 and 28 are both
printed, so we can see which private-state spaces the precondition
is in.   The node matching the conclusion of the rule is node 20.
The node actually inferred by the system is 38 -- the conclusion has
been placed in the private-state space \textit{writer negative
sentiment}.  Node 20 is also in the other private-state space, 
\textit{writer positive believesTrue}, but the system had already
inferred \textit{writer positive believesTrue Node 20}
by this point in the inference process.   In the body of this paper,
we are not showing the existing nodes that are inferred, but only the
newly created nodes (Appendix A shows inferences of both existing
and inferred nodes).  

Similarly, for the application of rule 7, node 20
matches the precondition, and 20 is in the same private-state spaces
as node 1.  Node 25 is the node matching the conclusion, and node 41
is the only new node created at this point;  the system had already
inferred \textit{writer positive believesTrue Node 25}.

Continuing with inference, since the writer has a negative sentiment toward the agent's positive
sentiment, the system infers that the writer disagrees with him and
thus that the writer is negative toward him:

\small
\textsf{ \\
\noindent
\textbf{\fbox{rule3.1}}   \\
\indent
S1 sentiment toward  
(S2 sentiment toward Z)  \\
\indent  $\Longrightarrow$ 
S1 agrees/disagrees with S2 that isGood/isBad Z \& 
S1 attitude toward Z   
\vspace{-1em}   \begin{multicols}{2}
\noindent
\textbf{Preconditions:} \\
\noindent
41 writer negative sentiment  \\
\indent 25 He positive sentiment  \\
\indent \indent 26 ideaOf \\
\indent \indent \indent  1 He trigger wars \\
\noindent
\vfill
\columnbreak
\noindent
\textbf{$\Longrightarrow$ Infer Node:} \\
\noindent
\noindent
50 writer disagrees with He that \\
\indent 49 isGood    \\
\indent \indent  26 ideaOf \\
\indent \indent \indent 1 He trigger wars \\
\noindent
\noindent
\end{multicols}   \vspace{-1em}
\noindent
\textbf{\fbox{rule4}}   \\
\indent
S1 agrees/disagrees with S2 that *  \\
\indent $\Longrightarrow$ 
S1 sentiment toward S2   
\vspace{-1em}   \begin{multicols}{2}
\noindent
\textbf{Preconditions:} \\
\noindent
50 writer disagrees with He that  \\
\indent 49 isGood    \\
\indent \indent 26 ideaOf \\
 \indent  \indent  \indent 1 He trigger wars \\
\vfill
\columnbreak
\noindent
\textbf{$\Longrightarrow$ Infer Node:} \\
\noindent
\noindent
55 writer negative sentiment    \\
\indent  3 He  \\
\end{multicols}  \vspace{-1em}
}
\normalsize

%

The MPQA annotators marked the writer's negative sentiment, choosing
the long spans \textit{is therefore \ldots industry} and
\textit{therefore planning \dots here and there} as attitude and
expressive subjective element spans, respectively.  They were not able
to pinpoint any clear sentiment phrases.  A machine learning system
trained on such examples would have difficulty recognizing the
sentiments.
The system, relying on the
negative connotation of \textit{war} and the gfbf information in the sentence,
is ultimately able to infer several sentiments, including the writer's negative sentiment toward the
\textit{trigger} event. 

\subsection{Inference toward the object of a gfbf}
\label{sentobject}
We now step through significant chains of reasoning, beginning
with perhaps the most basic inference: inferring an attitude
toward the object of a gfbf which is the target of a sentiment.
We will use an example given above:

\small
\textsf{ \\ 
\noindent
\noindent
``Is it no surprise then that MoveOn would attack Senator McCain.!?''   \\
\noindent
4 writer negative sentiment   \\
\indent 1 MoveOn attack Senator McCain  \\ 
}  \normalsize

\noindent 
Two rules infer the writer's sentiment toward Senator McCain, rule1 then rule2:

\small
\textsf{ \\ 
\noindent
\textbf{\fbox{rule1}}   \\
\indent S sentiment toward A goodFor/badFor T   \\
\indent $\Longrightarrow$
S sentiment toward the idea of A goodFor/badFor T  
\vspace{-1em}  \begin{multicols}{2}
\noindent
\textbf{Preconditions:} \\
\noindent
4 writer negative sentiment \\ 
\indent 1 MoveOn attack Senator McCain \\
\noindent
\vfill  \columnbreak
\noindent
\textbf{$\Longrightarrow$ Infer Node:} \\
\noindent
\noindent
139 writer negative sentiment  \\
\indent 138 ideaOf \\
\indent \indent 1 MoveOn attack Senator McCain  
\end{multicols}  \vspace{-1em} } \normalsize

\small
\textsf{ \\ 
\noindent
\noindent
\textbf{\fbox{rule2}}   \\
\indent S sentiment toward the idea of A goodFor/badFor T    \\
\indent $\Longrightarrow$
S sentiment toward T   
\vspace{-1em}  \begin{multicols}{2}
\noindent
\textbf{Preconditions:} \\
\noindent 139 writer negative sentiment \\
\indent 138 ideaOf \\
\indent \indent 1 MoveOn attack Senator McCain \\
\vfill  \columnbreak
\noindent
\textbf{$\Longrightarrow$ Infer Node:} \\
\noindent 
\noindent 140 writer positive sentiment  \\
\indent 2 Senator McCain \\
\end{multicols}  \vspace{-1em}
} \normalsize

\subsection{Inferences toward an animate agent of a gfbf}
\label{animateAgent}

Inference of an attitude toward
an animate agent of a gfbf proceeds by first inferring that the
action is intentional.
Continuing with our current example:

\small
\textsf{ \\ 
\noindent
\noindent
``Is it no surprise then that MoveOn would attack Senator McCain.!?''   \\
\noindent
\textbf{\fbox{rule6}}   \\
\indent A goodFor/badFor T, where A is animate    \\
\indent $\Longrightarrow$
A intended A goodFor/badFor T  
\vspace{-1em}  \begin{multicols}{2}
\noindent
\textbf{Preconditions:}   \\
\noindent \textit{4 writer negative sentiment} \\
\indent   1 MoveOn attack Senator McCain \\
\vfill  \columnbreak
\noindent
\textbf{$\Longrightarrow$ Infer Node:} \\
\noindent 
\noindent \textit{142 writer negative sentiment} \\
\indent 141 MoveOn positive intends \\
\indent \indent 1 MoveOn attack Senator McCain \\
\noindent
\noindent 
\noindent \textit{143 writer positive believesTrue} \\
\indent 141 MoveOn positive intends  \\
\indent \indent 1 MoveOn attack Senator McCain \\
\end{multicols}    \vspace{-1em}
}  \normalsize

\noindent
And then that MoveOn is positive toward their intentional event: 

\small
\textsf{ \\
\noindent
\textbf{\fbox{rule7}}   \\
\indent
S intended S goodFor/badFor T    \\
\indent  $\Longrightarrow$
S positive-sentiment toward ideaOf S goodFor/badFor T   
\vspace{-1em} \begin{multicols}{2}
\noindent
\textbf{Preconditions:} \\
\noindent  \textit{142 writer negative sentiment} \\
\indent   141 MoveOn positive intends \\
\indent \indent   1 MoveOn attack Senator McCain  \\
\noindent  \textit{143 writer positive believesTrue} \\
\indent   141 MoveOn positive intends \\
\indent \indent   1 MoveOn attack Senator McCain 
\vfill  \columnbreak
\noindent
\textbf{$\Longrightarrow$ Infer Node:}  \\
\noindent 
\noindent \textit{145 writer negative sentiment} \\
\indent 144 MoveOn positive sentiment \\
\indent \indent 138 ideaOf \\
\indent \indent \indent 1 MoveOn attack Senator McCain  \\
\noindent 
\noindent \textit{146 writer positive believesTrue} \\
\indent 144 MoveOn positive sentiment \\
\indent \indent 138 ideaOf  \\
\indent \indent \indent 1 MoveOn attack Senator McCain 
\end{multicols}   \vspace{-1em}
}  \normalsize

\noindent 
A fundamental rule now applies:  if the writer has a negative
sentiment toward MoveOn having a positive sentiment toward X,
{\it the writer disagrees with MoveOn} about X.  Further, by definition (of disagreement),
the writer has a negative sentiment toward X.

\small
\textsf{ \\ 
\noindent
\textbf{\fbox{rule3.1}}   \\
\noindent
\indent
S1 sentiment toward S2 sentiment toward Z  \\
\indent  $\Longrightarrow$
S1 agrees/disagrees with S2 that isGood/isBad Z \&
S1 sentiment toward Z  
\vspace{-1em}   \begin{multicols}{2}
\noindent
\textbf{Preconditions:}  \\
\noindent 145 writer negative sentiment \\
\indent 144 MoveOn positive sentiment \\
\indent \indent 138 ideaOf  \\
\indent \indent \indent 1 MoveOn attack Senator McCain \\
\vfill   \columnbreak
\noindent
\textbf{$\Longrightarrow$ Infer Node:} \\
\noindent 
\noindent 153 writer disagrees with MoveOn that \\
\indent 152 isGood \\
\indent \indent 138 ideaOf  \\
\indent \indent \indent 1 MoveOn attack Senator McCain \\
\noindent
\noindent 
\end{multicols}    \vspace{-1em} 
}  \normalsize  

\noindent
Finally, the system infers that the writer is negative toward MoveOn, {\it
  since the writer
disagrees with them:}

\small
\textsf{ \\ 
\noindent
\textbf{\fbox{rule4}}   \\
\indent
S1 agrees/disagrees with S2 that *    \\
\indent   $\Longrightarrow$
S1 sentiment toward S2   
\vspace{-1em}   \begin{multicols}{2}
\noindent
\textbf{Preconditions:}  \\
\noindent 153 writer disagrees with MoveOn that  \\
\indent 152 isGood  \\
\indent \indent 138 ideaOf  \\
\indent \indent \indent 1 MoveOn attack Senator McCain  \\
\vfill  \columnbreak
\noindent
\textbf{$\Longrightarrow$ Infer Node:}  \\
\noindent 
\noindent 158 writer negative sentiment \\
\indent 3 MoveOn \\
\end{multicols}   \vspace{-1em}
}  \normalsize

\subsection{Attitude Ascription}
\label{attitudeAscriptionEG}
The system ascribes its own reasoning to the sources in the sentence.
Above, the system inferred that the writer believes that MoveOn is
positive toward the idea of MoveOn attacking Senator McCain (node 146).  Now, it
applies rule2 {\it again} (note that the precondition, 144, is a child
of node 146):

\small
\textsf{ \\
\noindent
\noindent
``Is it no surprise then that MoveOn would attack Senator McCain.!?''   \\
\noindent
\textbf{\fbox{rule2}}   \\
\indent S sentiment toward the idea of A goodFor/badFor T    \\
\indent $\Longrightarrow$
S sentiment toward T 
\vspace{-1em} \begin{multicols}{2}
\noindent
\textbf{Preconditions:} \\
\noindent  \textit{145 writer negative sentiment} \\
\indent   144 MoveOn positive sentiment   \\
\indent \indent   138 ideaOf  \\
\indent \indent \indent   1 MoveOn attack Senator McCain  \\
\noindent  \textit{146 writer positive believesTrue}  \\
\indent   144 MoveOn positive sentiment   \\
\indent \indent   138 ideaOf  \\
\indent \indent \indent   1 MoveOn attack Senator McCain  
\vfill   \columnbreak
\noindent
\textbf{$\Longrightarrow$ Infer Node:} \\
\noindent 
\noindent \textit{150 writer negative sentiment}   \\
\indent 149 MoveOn negative sentiment   \\
\indent \indent 2 Senator McCain  \\
\noindent
\noindent 
\noindent \textit{151 writer positive believesTrue}  \\
\indent 149 MoveOn negative sentiment   \\
\indent \indent 2 Senator McCain 
\end{multicols} \vspace{-1em}
}  \normalsize

\noindent 
Recall that the system inferred above that {\bf the writer} is {\bf positive} toward Senator
McCain; here it just inferred that the writer believes that {\bf MoveOn} is
{\bf negative} toward him.\\

Moreover, the system inferred that the writer is negative that
MoveOn is negative toward Senator McCain (node 150).   Thus, the system
infers another disagreement:

\small
\textsf{ \\ 
\noindent
\noindent
\textbf{\fbox{rule3.1}}   \\
\indent
S1 sentiment toward S2 sentiment toward Z  \\
\indent  $\Longrightarrow$
S1 agrees/disagrees with S2 that isGood/isBad Z \&
S1 sentiment toward Z   
\vspace{-1em}    \begin{multicols}{2}
\noindent
\textbf{Preconditions:}  \\
\noindent 150 writer negative sentiment   \\
\indent 149 MoveOn negative sentiment   \\
\indent \indent 2 Senator McCain  \\
\vfill   \columnbreak
\noindent
\textbf{$\Longrightarrow$ Infer Node:}  \\
\noindent 
\noindent 156 writer disagrees with MoveOn that \\
\indent 155 isBad \\
\indent \indent 2 Senator McCain  \\
\noindent
\noindent 
\end{multicols}     \vspace{-1em}
}  \normalsize

\noindent
We won't show it here, but rule4 fires a second time, inferring node 158 again
(that the writer is negative toward MoveOn, since the writer disagrees
with him).

\subsection{Inferences toward an inanimate ({\it thing}) agent}
Consider the sentence \textit{Mother is upset that the tree fell on the
boy}.  Without evidence to the contrary, Mother is negative toward
the tree {\it because} it fell on the boy; before the incident, she
may have loved the tree.  She doesn't ``disagree'' or blame the tree,
since it isn't animate.  Thus, in these cases, the system requires
that the event actually happened, i.e., that the event is
\textit{substantial}. And, only one simple inference is made concerning the
agent (the tree), namely that the writer believes that Mother is
negative toward it.

Here is the input provided to the system:

\small
\textsf{ \\ 
\noindent
\noindent
``Mother is upset that the tree fell on the boy''   \\
E1 gfbf         $\langle$the tree:thing, badFor (fell on,fall on:lexEntry), the boy$\rangle$ \\
S1 subjectivity $\langle$mother, negative sentiment (upset), E1$\rangle$ \\
B1 privateState $\langle$writer, positive believesTrue (""), S1$\rangle$ \\
B2 privateState $\langle$writer, positive believesTrue (""), E1$\rangle$ \\
Prop1 p(B2,substantial)  \\ 
}  \normalsize

\noindent
Notice the  ``p=substantial'' on the last line (see \textit{believesTrue} in Section
\ref{kr}) and the indication that the agent is a thing (\textit{the
  tree:thing}). 
Here are the nodes built for the input:

\small
\textsf{ \\ 
\noindent
13 writer positive believesTrue substantial \\
\indent    6 the tree:thing fell on the boy \\
\noindent
11 writer positive believesTrue \\
\indent    9 mother negative sentiment \\ 
\indent \indent        6 the tree:thing fell on the boy \\
}  \normalsize 

\noindent
Note that we have only told the system that {\it the writer} believesTrue substantial
that the tree fell on the boy.  The rule involved, rule9, has an
assumption.  For this example, the assumption is that {\it Mother}
believesTrue substantial
that the tree fell on the boy.  Since the writer's
belief provides a basis for the assumption,
and since there
is no evidence to the contrary, the rule successfully fires.

\small
\textsf{ \\ 
\noindent
\noindent
\textbf{\fbox{rule9}}   \\
\indent
S sentiment toward A goodFor/badFor T, where A is a thing \&   \\
\indent
(Assume S positive believesTrue substantial) A goodFor/badFor T    \\
\indent   $\Longrightarrow$
S sentiment toward A   
 \vspace{-1em} \begin{multicols}{2}
\noindent
\textbf{Preconditions:} \\
\noindent
\textit{11 writer positive believesTrue}  \\
\indent   9 mother negative sentiment   \\
\indent \indent   6 the tree fell on the boy  \\
\textbf{Assumptions:}  \\
\noindent 178 mother positive believesTrue substantial  \\
\indent 6 the tree fell on the boy  
\vfill   \columnbreak
\noindent
\textbf{$\Longrightarrow$ Infer Node:}  \\
\noindent 
\noindent \textit{180 writer positive believesTrue}  \\
\indent \begin{scriptsize}178 mother positive believesTrue substantial\end{scriptsize} \\
\indent \indent 6 the tree fell on the boy  \\
\noindent
\noindent 
\noindent \textit{181 writer positive believesTrue}  \\
\indent 179 mother negative sentiment   \\
\indent \indent 8 the tree \\
\end{multicols}  \vspace{-1em}
}  \normalsize

\noindent
Other rules then fire for this example and the system infers, for example,
that mother is positive toward the boy (since she is upset that
something bad happened to him).  

This is a good place to introduce the
{\it by spaces} summary display of the nodes built for a sentence.
The \textit{by spaces} display consists of repetitions of the following.
One or more lines in square
brackets is printed; each of these is a private-state space.  Then, a node is
printed.  This means that that node is in all of those private-state
spaces. The numbers are the node numbers used elsewhere in the output.

Below is the \textit{by spaces} display for the current sentence.
The tree, for example, is only in one space:  writer positive
believesTrue (writer +B) mother has a negative sentiment (mother -S) (see node 181,
which we inferred just above).

The end of the display shows that node 6 is in three spaces, two from
the input, and the third from a rule application.  In particular, node
180 was built due to the assumption in rule9.

\small
\textsf{ \\ 
\noindent
\textit{[177 writer +B mother +S]}  \\
\noindent
7 the boy  \\
\noindent
\textit{[181 writer +B mother -S]}  \\
\noindent
8 the tree \\
\noindent
\textit{[175 writer +B mother -S]}  \\
\noindent
173 ideaOf  \\
\indent 6 the tree fell on the boy  \\
\noindent
\textit{From Input: [11 writer +B mother -S]}  \\
\noindent
\textit{From Input: [13 writer +B]}  \\
\noindent
\textit{[180 writer +B mother +B]}  \\
\noindent
6 the tree fell on the boy \\ 
}  \normalsize

\subsection{From sentiment toward the object of a gfbf to sentiment toward the
  gfbf itself}
\label{objectgfbf}

It is good if something bad happens to something bad, or if something
good happens to something good; it's bad if something bad happens to
something good, or if something good happens to something bad.
In our framework, ``something bad'' or ``something good'' is
manifested as sentiment toward that thing.  If that thing is the
object of a gfbf, we can infer a sentiment toward the gfbf
as well. 

We illustrate this inference with the following example:

\small
\textsf{ \\ 
\noindent
\noindent
``Obama will finally bring skyrocketing health care costs under control''   \\ 
E1 gfbf         $\langle$Obama, badFor (bring-under-control,bringUnderControl:lexEntry), skyrocketing-health-care-costs$\rangle$ \\
B1 privateState $\langle$writer, positive believesTrue (""), E1$\rangle$ \\
S1 subjectivity $\langle$writer, negative sentiment (skyrocketing), skyrocketing-health-care-costs$\rangle$ \\ \\ 
\noindent
17 writer positive believesTrue \\
\indent 14 Obama bring-under-control skyrocketing-health-care-costs \\
\noindent
19 writer negative sentiment \\
\indent 15 skyrocketing-health-care-costs \\ 
}  \normalsize

\noindent
\textit{Skyrocketing} is a clear negative evaluation of the object of the
gfbf.  Bringing costs under control is badFor the costs.
Thus, the system infers that the writer is {\bf positive} toward the
gfbf.

\small
\textsf{ \\ 
\noindent
\noindent
\textbf{\fbox{rule8}}   \\
\indent
S positive believesTrue A goodFor/badFor T \& S sentiment toward T    \\
\indent   $\Longrightarrow$
S sentiment toward A goodFor/badFor T   
\vspace{-1em}   \begin{multicols}{2}
\noindent
\textbf{Preconditions:}  \\
\noindent 17 writer positive believesTrue   \\
\indent 14 Obama bring-under-control costs\footnote{We use \textit{costs} to represent \textit{skyrocketing-health-care-costs} for short in node 14.}  \\
\noindent 19 writer negative sentiment \\
\indent 15 skyrocketing-health-care-costs  \\
\vfill   \columnbreak  
\noindent
\textbf{$\Longrightarrow$ Infer Node:}  \\
\noindent 
\noindent 194 writer positive sentiment  \\
\indent 14 Obama bring-under-control costs \\ 
\end{multicols}    \vspace{-1em}
}  \normalsize

\noindent
Now that the system has inferred a sentiment toward a gfbf,
inference proceeds from here as it does from the {\bf start} for the first
example above, \textit{Is it no surprise then that MoveOn would attack
Senator McCain.!?}

\subsection{Complex Attitude Ascription}
\label{complexAttitudeAscription}
This section shows nodes built for a sentence with nested
sentiment.
First, here are the inputs and the system's display of the graph built
from the input.

\small
\textsf{ \\ 
\noindent
\noindent
``MoveOn is livid that the Republicans keep opposing
Obama's efforts to raise taxes on the rich''   \\
E1 gfbf         $\langle$Obama, goodFor (raise,raise:lexEntry), taxes on the rich$\rangle$ \\
S1 subjectivity $\langle$the Republicans, negative sentiment (opposing), E1$\rangle$ \\
S2 subjectivity $\langle$MoveOn, negative sentiment (livid), S1$\rangle$ \\
B1 privateState $\langle$writer, positive believesTrue (""), S2$\rangle$ \\ 
}
\textsf{ \\
27 writer positive believesTrue   \\
\indent 25 MoveOn negative sentiment   \\
\indent \indent 23 the Republicans negative sentiment   \\
\indent \indent \indent 20 Obama raising taxes on the rich  \\ 
}  \normalsize

\noindent
Now, here is the ``by spaces'' representation of several of the nodes built
for this sentence (see Appendix A for the full listing):

\small
\textsf{ \\ 
\noindent \textit{[315 writer +B]}  \\
\noindent
313 MoveOn disagrees with the Republicans that \\
\indent 312 isBad \\
\indent \indent 281 Obama positive sentiment \\
\indent \indent \indent 21 taxes on the rich  \\
\noindent
\noindent \textit{[370 writer +B]}  \\
\noindent
368 MoveOn disagrees with the Republicans that \\
\indent 367 isBad \\
\indent \indent 22 Obama  \\
\noindent \textit{[321 writer +B MoveOn -S]}  \\
\noindent \textit{[323 writer +B MoveOn +B]}  \\
\noindent
318 the Republicans disagrees with Obama that \\
\indent 317 isGood \\
\indent \indent 21 taxes on the rich  \\
\noindent
\noindent \textit{[366 writer +B]}  \\
\noindent
364 MoveOn agrees with Obama that \\
\indent 317 isGood \\
\indent \indent 21 taxes on the rich  \\
\noindent \textit{[235 writer +B MoveOn -S the Republicans -S]}  \\
\noindent \textit{[236 writer +B MoveOn +B the Republicans -S]}  \\
\noindent \textit{[249 writer +B MoveOn +S]}  \\
\noindent \textit{[271 writer +B MoveOn -S the Republicans -S Obama +S]}  \\
\noindent \textit{[274 writer +B MoveOn +B the Republicans -S Obama +S]}  \\
\noindent \textit{[277 writer +B MoveOn +B the Republicans +B Obama +S]}  \\
\noindent \textit{[304 writer +B MoveOn +S Obama +S]}  \\
\noindent \textit{[337 writer +B MoveOn +B Obama +S]}  \\
\noindent
232 ideaOf \\
\indent 20 Obama raise taxes on the rich  \\
\noindent
\noindent \textit{[256 writer +B MoveOn -S]}  \\
\noindent
24 the Republicans \\
\noindent
\noindent \textit{[330 writer +B MoveOn -S the Republicans -S]}  \\
\noindent \textit{[332 writer +B MoveOn +B the Republicans -S]}  \\
\noindent \textit{[371 writer +B MoveOn +S]}  \\
\noindent
22 Obama \\
\noindent
\noindent \textit{[242 writer +B MoveOn -S the Republicans -S]}  \\
\noindent \textit{[244 writer +B MoveOn +B the Republicans -S]}  \\
\noindent \textit{[254 writer +B MoveOn +S]}  \\
\noindent \textit{[284 writer +B MoveOn -S the Republicans -S Obama +S]}  \\
\noindent \textit{[287 writer +B MoveOn +B the Republicans -S Obama +S]}  \\
\noindent \textit{[290 writer +B MoveOn +B the Republicans +B Obama +S]}  \\
\noindent \textit{[316 writer +B MoveOn +S Obama +S]}  \\
\noindent \textit{[346 writer +B MoveOn +B Obama +S]}  \\
\noindent
21 taxes on the rich  \\
}  \normalsize

\subsection{Rules that Fire Only on Input}
\subsubsection{From connotation to sentiment}
\label{connToSent}
Rule8 (which infers sentiment toward a gfbf from sentiment toward its
object)
fires only if there is {\bf sentiment} toward the
object of a gfbf which, as with all attitudes, is attributed
to a source.  What if there is merely connotation in place of the
sentiment, for example, \textit{The attack is a fight
against justice} or \textit{He started a war} ({\it justice} has positive
connotation while {\it war} has negative connotation).
The way to handle this is to infer sentiment from connotation.
As with all the rules, an
explicit sentiment to the contrary would block the
inference of a sentiment from a connotation.
For example, the writer might be {\it glad} that
someone started a war, if he is an arms dealer (even though {\it war} has negative connotation,
and starting a war is goodFor it).
This section illustrates the inference from connotation to
sentiment, using the following sentence.

\small
\textsf{ \\ 
\noindent
\noindent
``GOP Attack on Health Care Reform Is a Fight Against Racial Justice.''   \\
E1 gfbf         $\langle$GOP, badFor (attack), Health-Care-Reform$\rangle$ \\
B1 privateState $\langle$writer, positive believesTrue (""), E1$\rangle$ \\
E2 gfbf         $\langle$GOP-Attack, badFor (fight), racial-justice$\rangle$ \\
B2 privateState $\langle$writer, positive believesTrue (""), E2$\rangle$ \\ \\
\noindent
32 writer positive believesTrue   \\
\indent 29 GOP attack Health-Care-Reform   \\
\noindent
37 writer positive believesTrue   \\
\indent 34 GOP-Attack fight racial-justice   \\
}  \normalsize

\noindent
This sentence has two gfbfs: the GOP attacking health care reform
and the GOP's attack fighting racial justice.
The lexicon has an entry for \textit{justice} indicating that it has positive
connotation.  However, it does not have an entry for \textit{health care
reform} because, in our healthcare dataset (\cite{dengacl2013}), both negative and positive attitudes
toward reform are common. 

Note that a rule fires on {\it all} possible matches.  So, in the full
output (Appendix A) you will
see that the system makes inferences for both gfbfs.
The connotation rule, rule10, only matches the object of
the second gfbf, since only its text anchor has a connotation in
the lexicon. 

Note that rule10 may only infer sentiments of the writer -- not other
sources in the sentence.  The reason is that it is \textit{the
writer's} words that appear in the sentence.  (At least, the
current system assumes that all words in the sentence are the writer's
words.  In the future, a component should be added that recognizes
quotations and rule10 should be modified appropriately.)

\textsf{ \\ 
\noindent
\noindent
\textbf{\fbox{rule10}}   \\
\indent
(Assume Writer positive believesTrue) A goodFor/badFor T \&   \\
\indent
T's anchor is in connotation lexicon     \\
\indent     $\Longrightarrow$
Writer sentiment toward T   
\vspace{-1em}    \begin{multicols}{2}
\noindent
\textbf{Assumptions:}   \\
\noindent
32 writer positive believesTrue   \\
\indent 29 GOP-Attack fight racial-justice   \\
\vfill   \columnbreak
\noindent
\textbf{$\Longrightarrow$ Infer Node:}   \\
\noindent
\noindent
\noindent
\noindent
418 writer positive sentiment   \\
\indent 34 racial-justice  
\end{multicols}    \vspace{-1em}
}  \normalsize

\noindent
Now that this sentiment (node 418) has been inferred, inference proceeds as in
\ref{objectgfbf} for the sentence \textit{Obama will finally bring
skyrocketing health care costs under control.}  Following is a subset of the \textit{by
spaces} display of the nodes built for the sentence.

\small
\textsf{ \\ 
\noindent
454 writer disagrees with GOP-Attack that   \\
\indent 453 isBad   \\
\indent \indent 34 racial-justice   \\
\noindent
``GOP Attack on Health Care Reform Is a Fight Against Racial Justice.''   \\
\noindent
\textit{[418 writer +S]}   \\
\noindent
\textit{[435 writer +B GOP-Attack -S]}   \\
\noindent
\textit{[448 writer -S GOP-Attack -S]}   \\
\noindent 34 racial-justice  \\ 
}  \normalsize

\noindent
The inferences often bring into sharp relief the fact that the words are the
writer's words and not those of the other sources in the sentence.
It is doubtful that the GOP would agree that it has a negative
sentiment toward
\textit{racial justice.}

\subsubsection{From sentiment toward the source to sentiment
  toward the private state}
\label{sourceattitude}

In this section, we return to the example we started in
Section \ref{datastructure}.

\small
\textsf{ \\ 
\noindent
\noindent
``Mayor Blooming idiot urges Congress to vote for gun control.''   \\
E1 gfbf        $\langle$Congress, goodFor (voting for,voteFor:lexEntry), gun control$\rangle$ \\
S1 subjectivity $\langle$Mayor-Blooming-idiot, positive sentiment (urges), E1$\rangle$ \\
B1 privateState $\langle$writer, positive believesTrue (""), S1$\rangle$ \\
S2 subjectivity $\langle$writer, negative sentiment (blooming idiot), Mayor-Blooming-idiot$\rangle$  \\ \\ 
\noindent
43 writer positive believesTrue   \\
\indent 41 Mayor-Blooming-idiot positive sentiment   \\
\indent \indent 38 Congress voting for gun control   \\
\noindent
45 writer negative sentiment   \\
\indent 42 Mayor-Blooming-idiot  \\ 
}  \normalsize

\noindent
In this sentence, the writer opts to explicitly express his negative sentiment
toward Mayor Bloomberg (\textit{Blooming idiot}), not toward Mayor Bloomberg's
attitude toward the gfbf (i.e., his positive attitude toward Congress doing something goodFor gun control).   Rule
rule5source infers a sentiment toward an attitude from a sentiment
toward the source of that attitude:

\small
\textsf{ \\ 
\noindent
\noindent
\textbf{\fbox{rule5source}}   \\
\indent
S1 sentiment toward S2 in the input \&   \\
\indent
(Assume S1 positive believesTrue) S2 privateState toward Z in input    \\
\indent   $\Longrightarrow$
S1 sentiment toward  S2 privateState toward Z   
\vspace{-1em}   \begin{multicols}{2}
\noindent
\textbf{Preconditions:}  \\
\noindent 45 writer negative sentiment \\
\indent 42 Mayor-Blooming-idiot \\
\noindent
\textbf{Assumptions:}  \\
\noindent 43 writer positive believesTrue \\
\indent \begin{scriptsize}41 Mayor-Blooming-idiot positive sentiment\end{scriptsize} \\
\indent \indent 38 Congress voting for gun control  \\
\vfill   \columnbreak
\noindent
\textbf{$\Longrightarrow$ Infer Node:}  \\
\noindent 
\noindent 444 writer negative sentiment \\
\indent \begin{scriptsize}41 Mayor-Blooming-idiot positive sentiment\end{scriptsize} \\
\indent \indent 38 Congress voting for gun control  \\
\noindent
\noindent 
\end{multicols}      \vspace{-1em}
}  \normalsize

\noindent
From this point, inference proceeds as it did for the example in \ref{complexAttitudeAscription},
\textit{MoveOn is livid that the Republicans keep opposing Obama's efforts
to raise taxes on the rich.}

This rule (rule5source) is not a general conceptual rule such as the
ones we've seen so far.   It is a rule concerning which explicit
sentiment expressions writers choose to include in a sentence.  Thus, the rule
applies only to nodes built to represent the input.
\footnote{As indicated in a footnote above, because the rule5 rules are not general inference rules, currently the system
will only allow the rule to fire once on a given precondition, even if
two assumptions are possible with the same precondition.   The idea is
that one fire gives us an explanation or reason for the
precondition. This option can be changed by flipping
a variable value from 1 to 0. 
}

Note that rule5source includes an assumption.  Here is an example for
which the assumption comes into play.

\small
\textsf{ \\ 
\noindent
\noindent
``Republicans oppose Obama because he supports the states legalizing gay marriage.''   \\
\noindent
\noindent E1 gfbf         $\langle$the states, goodFor (legalizing,legalize:lexEntry), gay marriage$\rangle$ \\
\noindent S1 subjectivity $\langle$Obama, positive sentiment (supports), E1$\rangle$ \\
\noindent B1 privateState $\langle$writer, positive believesTrue (""), S1$\rangle$ \\
\noindent S2 subjectivity $\langle$Republicans, negative sentiment (supports), Obama$\rangle$  \\
\noindent B2 privateState $\langle$writer, positive believesTrue (""), S2$\rangle$  \\ \\ 
\noindent 
55 writer positive believesTrue   \\
\indent 53 Republicans negative sentiment   \\
\indent \indent 50 Obama   \\
\noindent
51 writer positive believesTrue   \\
\indent 49 Obama positive sentiment   \\
\indent \indent 46 the states legalizing gay marriage  \\ 
}  \normalsize

\noindent
The preconditions of rule5source have to be ``at the same level'' in terms of
private-state spaces.  Via the assumption, the system ascribes to the
Republicans the writer's belief (node 51) that Obama
has a positive sentiment toward the gfbf (legalizing gay marriage).

\small
\textsf{ \\ 
\noindent
\noindent
\textbf{\fbox{rule5source}}   \\
\indent
S1 sentiment toward S2 in the input \&   \\
\indent
(Assume S1 positive believesTrue) S2 privateState toward Z in input    \\
\indent  $\Longrightarrow$
S1 sentiment toward   
  S2 privateState toward Z   
 \vspace{-1em} \begin{multicols}{2}
\noindent
\textbf{Preconditions:}  \\
\noindent \textit{55 writer positive believesTrue} \\
\indent   53 Republicans negative sentiment \\
\indent \indent   50 Obama \\
\textbf{Assumptions:}  \\
\noindent 592 Republicans positive believesTrue \\
\indent 49 Obama positive sentiment \\
\indent \indent 46 the states legalizing gay marriage \\
\vfill \columnbreak
\noindent
\textbf{$\Longrightarrow$ Infer Node:}   \\
\noindent 
\noindent \textit{594 writer positive believesTrue} \\
\indent 592 Republicans positive believesTrue \\
\indent \indent 49 Obama positive sentiment \\
\indent \indent \indent \begin{scriptsize}46 the states legalizing gay marriage\end{scriptsize} \\
\noindent
\noindent 
\noindent \textit{595 writer positive believesTrue} \\
\indent 593 Republicans negative sentiment \\
\indent \indent 49 Obama positive sentiment \\
\indent \indent \indent \begin{scriptsize}46 the states legalizing gay marriage\end{scriptsize}  \\ 
\end{multicols}    \vspace{-1em}
}  \normalsize

\noindent
Following is an example with a source other than the writer for which
the belief is explicitly in the sentence, so a new assumption does not
need to be made.

\small
\textsf{ \\ 
\noindent
\noindent
``Muslims also hate Obama because they think he 
supported the Koran burning event by Pastor Terry Jones in Florida on March 2011.''   \\
\noindent
\noindent S1 subjectivity $\langle$Muslims, negative sentiment (hate), Obama$\rangle$ \\
\noindent B1 privateState $\langle$writer, positive believesTrue (""), S1$\rangle$ \\
\noindent E1 gfbf         $\langle$Pastor-Terry-Jones, badFor (burning,burn:lexEntry), Koran$\rangle$ \\
\noindent S2 subjectivity $\langle$Obama, positive sentiment (supports), E1$\rangle$ \\
\noindent B2 privateState $\langle$Muslims, positive believesTrue (""), S2$\rangle$ \\
\noindent B3 privateState $\langle$writer, positive believesTrue (""), B2$\rangle$ \\ \\ 
\noindent
66 writer positive believesTrue   \\
\indent 65 Muslims positive believesTrue   \\
\indent \indent 64 Obama positive sentiment   \\
\indent \indent \indent 61 Pastor-Terry-Jones burning Koran   \\
\noindent
59 writer positive believesTrue   \\
\indent 56 Muslims negative sentiment   \\
\indent \indent 57 Obama   \\ \\
\noindent
\textbf{\fbox{rule5source}}   \\
\indent
S1 sentiment toward S2 in the input \&   \\
\indent
(Assume S1 positive believesTrue) S2 privateState toward Z in input    \\
\indent    $\Longrightarrow$
S1 sentiment toward   
  S2 privateState toward Z   
 \vspace{-1em} \begin{multicols}{2}
\noindent
\textbf{Preconditions:}  \\
\noindent
\textit{59 writer positive believesTrue} \\
\indent   56 Muslims negative sentiment \\
\indent \indent   57 Obama \\
\noindent
\textbf{Assumptions:}  \\
\noindent  \textit{66 writer positive believesTrue} \\
\indent   65 Muslims positive believesTrue \\
\indent \indent   64 Obama positive sentiment \\
\indent \indent \indent   \begin{scriptsize}61 Pastor-Terry-Jones burning Koran\end{scriptsize}
\vfill   \columnbreak
\noindent
\textbf{$\Longrightarrow$ Infer Node:} \\
\noindent 
\noindent \textit{791 writer positive believesTrue} \\
\indent 790 Muslims negative sentiment \\
\indent \indent 64 Obama positive sentiment \\
\indent \indent \indent \begin{scriptsize}61 Pastor-Terry-Jones burning Koran\end{scriptsize}  \\
\noindent
\noindent 
\end{multicols}    \vspace{-1em}
}  \normalsize

\subsubsection{From sentiment toward the agent of a gfbf to
  sentiment toward the gfbf}
\label{agentgfbf}

This rule is very similar to the one just above.  It infers a
sentiment toward a gfbf from a sentiment toward the agent of the
event. This rule also applies only to nodes built from the input, and
includes the same type of assumption as rule5source does.  Thus, we
only show one example here.

\small
\textsf{ \\ 
\noindent
\noindent
``Muslims started hating Obama when he ordered  
   the US Army to kill Osama bin Laden''  \\
\noindent
\noindent S1 subjectivity $\langle$Muslims, negative sentiment (hate), Obama$\rangle$ \\
\noindent B1 privateState $\langle$writer, positive believesTrue (""), S1$\rangle$ \\
\noindent E1 gfbf         $\langle$US Army, badFor (kill,kill:lexEntry), Osama bin Laden$\rangle$ \\
\noindent I1 influencer   $\langle$Obama, retain (ordered,order:lexEntry), E1$\rangle$ \\
\noindent B2 privateState $\langle$writer, positive believesTrue (""), I1$\rangle$ \\ \\ 
 \noindent 76 writer positive believesTrue \\ 
\indent   75 Obama $\langle$retain$\rangle$ \\
\indent \indent   72 US Army kill Osama bin Laden \\
\noindent 70 writer positive believesTrue \\
 \indent   67 Muslims negative sentiment  \\
\indent \indent        68 Obama \\
\noindent
\noindent
\textbf{influencer node}:  \\
 \noindent \textit{76 writer positive believesTrue} \\ 
\indent   75 Obama $\langle$retain$\rangle$ \\
\indent \indent   72 US Army kill Osama bin Laden \\
\noindent
\textbf{new gfbf}:  \\
\noindent  \textit{976 writer positive believesTrue}   \\
\indent   975 Obama badFor Osama bin Laden \\ \\
\noindent
\textbf{\fbox{rule5agent}}   \\
\indent
S1 sentiment toward A in input \&   \\
\indent
(Assume S1 sentiment toward) A goodFor/badFor T in input    \\
\indent  $\Longrightarrow$
S1 sentiment toward A goodFor/badFor T   
\vspace{-1em} \begin{multicols}{2}
\noindent
\textbf{Preconditions:} \\
\noindent \textit{70 writer positive believesTrue}  \\
\indent   67 Muslims negative sentiment \\
\indent \indent   68 Obama  \\
\noindent
\textbf{Assumptions:} \\
\noindent 977 Muslims negative sentiment \\
\indent 975 Obama badFor Osama bin Laden  \\
\vfill    \columnbreak
\noindent
\textbf{$\Longrightarrow$ Infer Node:} \\
\noindent 
\noindent \textit{978 writer positive believesTrue} \\
\indent 977 Muslims negative sentiment  \\
\indent \indent 975 Obama badFor Osama bin Laden  \\
\noindent
\noindent 
\end{multicols}   \vspace{-1em}
}  \normalsize

\noindent
This completes the section on rules that fire only on input nodes.
Looking to the future, we believe that the rule5* rules might be a good place
to start when extending the approach to the level of the discourse.
Often, there are patterns such as \textit{The Muslims hate Obama. He did
X, he supports Y, etc.}

\subsection{Arguing Subjectivity}
\label{arguingSubjectivity}
So far in this document, we've considered {\it sentiment subjectivity}:
positive and negative emotions, evaluations, and judgments.  Another
important type of subjectivity is {\it arguing subjectivity}, where
someone argues (1) that something is or isn't true, or
(2) that something should or should not be done.   

The first type of arguing subjectivity is given in the input as subjectivity
where the attitude type of the private state is believesTrue and, if the
target of the private state is a gfbf, the attribute p is given the
value substantial.
That is, the interpretation supported under the
framework is that, if someone arguesTrue an event, then what they are
arguing is that the event actually occurred (or will occur).  (Down
the road, I foresee the NLP system's task as recognizing arguing
subjectivity based on lexical items such as ``accuse''; the
{\it p=substantial} attribute would simply be added automatically whenever arguing
subjectivity is placed on the input.)

The
second is subjectivity where the attitude type of the private state
is believesShould.

To date, we haven't addressed the second type.  But, the system is able
to make an interesting inference with respect to the first type.

Consider the following example (we will consider the crossed-out lines
below).

\small
\textsf{ \\ 
\noindent
\noindent
``Republicans roared onto the post-State-of-the-Union morning shows  
accusing President Obama of waging class warfare against the rich''   \\
\noindent
\noindent E1 gfbf         $\langle$obama, badFor (waging class warfare against,wagingClassWarfare:lexEntry), the rich$\rangle$  \\
\noindent B1 subjectivity $\langle$republicans, positive believesTrue (accusing), E1$\rangle$  \\
\noindent B2 privateState $\langle$writer, positive believesTrue (""), B1$\rangle$  \\
\noindent Prop1 p(B1,substantial) \\
\noindent S1 subjectivity $\langle$writer, negative sentiment (roared), republicans$\rangle$  \\
\noindent \sout{S2 subjectivity $\langle$republicans, negative sentiment (accusing), obama$\rangle$} \\
\noindent \sout{B3 privateState $\langle$writer, positive believesTrue (""), S2$\rangle$}  \\ \\ 
\noindent
82 writer positive believesTrue   \\
\indent 80 republicans positive believesTrue substantial   \\
\indent \indent 77 obama waging class warfare against the rich   \\
\noindent
84 writer negative sentiment   \\
\indent 81 republicans  \\ 
}  \normalsize

\noindent
Inference begins with an application of rule5source:

\small
\textsf{ \\ 
\noindent
\noindent
\textbf{\fbox{rule5source}}   \\
\indent
S1 sentiment toward S2 in the input \&   \\
\indent
(Assume S1 positive believesTrue) S2 privateState toward Z in input    \\
\indent    $\Longrightarrow$
S1 sentiment toward   
  S2 privateState toward Z  \\ 
}  \normalsize

\noindent
Resulting in this attitude of the writer's:

\small
\textsf{ \\ 
\noindent 1027 writer negative sentiment \\
\indent 80 republicans positive believesTrue substantial \\
\indent \indent 77 obama waging class warfare against the rich  \\ 
}  \normalsize

\noindent
In the examples so far, the system has inferred positive and negative
sentiments (toward gfbfs and their agents and objects).   However,
{\it the writer isn't negative due to sentiment}; {\bf he's negative because
he doesn't think it's true!}

\small
\textsf{ \\ 
\noindent
\noindent
\textbf{\fbox{rule3.2}}   \\
\indent
S1 sentiment toward
  S2 positive/negative believesTrue substantial Z \\
\indent $\Longrightarrow$
S1 agrees/disagrees with S2 that isTrue/isFalse Z \& \\
\indent \indent S1 pos/neg believesTrue substantial Z
\vspace{-1em}   \begin{multicols}{2}   
\noindent
\textbf{Preconditions:}  \\
\noindent 1027 writer negative sentiment  \\
\indent \begin{scriptsize}80 republicans positive believesTrue substantial\end{scriptsize}   \\
\indent \indent \begin{scriptsize}77 obama waging class warfare against rich\end{scriptsize}  \\
\vfill   \columnbreak
\noindent
\textbf{$\Longrightarrow$ Infer Node:} \\
\noindent 
\noindent 1029 writer disagrees with republicans that  \\
\indent 1028 isTrue   \\
\indent \indent \begin{scriptsize}77 obama waging class warfare against rich\end{scriptsize}  \\
\noindent
\noindent 1030 writer negative believesTrue substantial  \\
\indent  \begin{scriptsize}77 obama waging class warfare against rich\end{scriptsize} \\ 
\end{multicols}    \vspace{-1em}
}  \normalsize

Let's step through the rest of the inferences for this sentence, to
see how things fit together (some comments have been added).

\small
\textsf{ \\ 
\noindent
\noindent
\textbf{\fbox{rule4}}   \\
\indent
S1 agrees/disagrees with S2 that *    \\
\indent   $\Longrightarrow$
S1 sentiment toward S2   
\vspace{-1em}   \begin{multicols}{2}
\noindent
\textbf{Preconditions:}  \\
\noindent 1029 writer disagrees with republicans that  \\
\indent 1028 isTrue  \\
\indent \indent \begin{scriptsize}77 obama waging class warfare against rich\end{scriptsize}  \\
\vfill   \columnbreak
\noindent
\textbf{$\Longrightarrow$ Infer Node:} \\
\noindent Existing: \\
\noindent 84 writer negative sentiment  \\
\indent 81 republicans \\ 
\end{multicols}    \vspace{-1em}
}  \normalsize

\noindent
The system infers that the writer is negative toward the republicans
because he disagrees with them about something.  This matches the
negative subjectivity in the input.\footnote{The system often infers
nodes that already exist.  On the full printout (see Appendix A),
inferences of existing nodes are included, marked by ``Existing:''
as above for Node 84.}

Continuing on, the system infers an intention and then the agent's
positive attitude toward his intended action.  Note that node 1030 is
a negative believesTrue private state.  Negative beliefs block
space-extension inferences.  Thus, the system does not infer that the
writer believes that Obama intends the action.

\small
\textsf{ \\ 
\noindent
\noindent
\textbf{\fbox{rule6}}   \\
\indent A goodFor/badFor T, where A is animate    \\
\indent $\Longrightarrow$
A intended A goodFor/badFor T   
\vspace{-1em}    \begin{multicols}{2}
\noindent
\textbf{Preconditions:}  \\
\noindent  \textit{1030 writer negative believesTrue substantial}  \\
\indent   \begin{scriptsize}77 obama waging class warfare against rich\end{scriptsize} \\
\noindent  \textit{82 writer positive believesTrue}   \\
\indent   \begin{scriptsize}\textit{80 republicans positive believesTrue substantial}\end{scriptsize}  \\
\indent \indent   \begin{scriptsize}77 obama waging class warfare against rich\end{scriptsize} \\
\noindent  \textit{1027 writer negative sentiment}    \\
\indent   \begin{scriptsize}\textit{80 republicans positive believesTrue substantial}\end{scriptsize}  \\
\indent \indent   \begin{scriptsize}77 obama waging class warfare against rich\end{scriptsize} 
\vfill   \columnbreak
\noindent
\textbf{$\Longrightarrow$ Infer Node:} \\
\noindent  
\noindent  \textit{1034 writer positive believesTrue}   \\
\indent \textit{1033 republicans positive believesTrue}   \\
\indent \indent 1032 obama positive intends   \\
\indent \indent \indent \begin{scriptsize}77 obama waging class warfare ...\end{scriptsize} \\
\noindent
\noindent 
\noindent \textit{1036 writer negative sentiment}    \\
\indent \textit{1033 republicans positive believesTrue}   \\
\indent \indent 1032 obama positive intends   \\
\indent \indent \indent \begin{scriptsize}77 obama waging class warfare ... \end{scriptsize} \\
\noindent
\noindent Inference blocked in space [writer -B]: \\
\noindent because it contains a negative believesTrue. \\
\end{multicols}  
\noindent
\textbf{\fbox{rule7}}   \\
\indent
S intended S goodFor/badFor T    \\
\indent   $\Longrightarrow$
S positive-sentiment toward ideaOf S goodFor/badFor T   
\vspace{-1em} \begin{multicols}{2}
\noindent
\textbf{Preconditions:}  \\
\noindent  \textit{1034 writer positive believesTrue}   \\
\indent  \textit{1033 republicans positive believesTrue}   \\
\indent \indent   1032 obama positive intends   \\
\indent \indent \indent   77 obama waging class warfare ... \\
\noindent  \textit{1036 writer negative sentiment}    \\
\indent   \textit{1033 republicans positive believesTrue}   \\
\indent \indent   1032 obama positive intends   \\
\indent \indent \indent   77 obama waging class warfare ... \\
\vfill   \columnbreak
\noindent
\textbf{$\Longrightarrow$ Infer Node:}  \\
\noindent 
\noindent \textit{1040 writer positive believesTrue}   \\
\indent \textit{1039 republicans positive believesTrue}   \\
\indent \indent 1037 obama positive sentiment    \\
\indent \indent \indent 1038 ideaOf  \\
\indent \indent \indent \indent \begin{scriptsize}77 obama waging class warfare ...\end{scriptsize} \\
\noindent
\noindent 
\noindent \textit{1042 writer negative sentiment}    \\
\indent \textit{1039 republicans positive believesTrue}   \\
\indent \indent 1037 obama positive sentiment    \\
\indent \indent \indent 1038 ideaOf  \\
\indent \indent \indent \indent \begin{scriptsize}77 obama waging class warfare ...\end{scriptsize} \\
\end{multicols}    \vspace{-1em} 
}  \normalsize

\noindent
The system has just inferred (within the
spaces [writer +B republicans +B] and [writer -S republications +B]) that Obama has a positive sentiment toward
the idea of the gfbf.  Thus, the system infers he has a negative
sentiment toward the object of the gfbf (the rich):

\small
\textsf{ \\ 
\noindent
\noindent
\textbf{\fbox{rule2}}   \\
\indent S sentiment toward the idea of A goodFor/badFor T    \\
\indent $\Longrightarrow$
 S sentiment toward T   
\vspace{-1em}   \begin{multicols}{2}
\noindent
\textbf{Preconditions:}   \\
\noindent  \textit{1040 writer positive believesTrue}  \\
\indent   \textit{1039 republicans positive believesTrue}  \\
\indent \indent   1037 obama positive sentiment  \\
\indent \indent \indent   1038 ideaOf  \\
\indent \indent \indent \indent   \begin{scriptsize}77 obama waging class warfare... \end{scriptsize} \\
\noindent \textit{1042 writer negative sentiment}  \\
\indent   \textit{1039 republicans positive believesTrue}  \\
\indent \indent   1037 obama positive sentiment  \\
\indent \indent \indent   1038 ideaOf  \\
\indent \indent \indent \indent   \begin{scriptsize}77 obama waging class warfare... \end{scriptsize} \\
\vfill   \columnbreak
\noindent
\textbf{$\Longrightarrow$ Infer Node:}  \\
\noindent 
\noindent \textit{1045 writer positive believesTrue}  \\
\indent \textit{1044 republicans positive believesTrue}  \\
\indent \indent 1043 obama negative sentiment  \\
\indent \indent \indent 78 the rich  \\
\noindent
\noindent 
\noindent \textit{1047 writer negative sentiment}  \\
\indent \textit{1044 republicans positive believesTrue}  \\
\indent \indent 1043 obama negative sentiment  \\
\indent \indent \indent 78 the rich \\ 
\end{multicols}     \vspace{-1em}
}  \normalsize

\noindent
That is the end of the inferences for this sentence.

For the sake of comparison, consider
the sentence above,
\textit{Mayor Blooming idiot urges Congress to vote for gun control.}  The
\textit{Blooming idiot} sentence
has the same structure as the one we just looked at, except it has sentiment subjectivity
(\textit{urges}) where the current sentence has arguing subjectivity
(\textit{accusing}).
A segment of the \textit{by spaces} representation will remind you of the
inferences made for that sentence (note that \textit{gun control} is the
object of the gfbf):

\small
\textsf{ \\ 
\noindent
\noindent
``Mayor Blooming idiot urges Congress to vote for gun control.''   \\
\noindent \textit{[450 writer +B Mayor-Blooming-idiot +S]} \\
\noindent \textit{[451 writer -S Mayor-Blooming-idiot +S]} \\
\noindent \textit{[457 writer -S]}  \\
\noindent \textit{[479 writer +B Mayor-Blooming-idiot +S Congress +S]}  \\
\noindent \textit{[481 writer -S Mayor-Blooming-idiot +S Congress +S]}  \\
\noindent \textit{[483 writer +B Mayor-Blooming-idiot +B Congress +S]}  \\
\noindent \textit{[504 writer -S Congress +S]}  \\
\noindent \textit{[522 writer +B Congress +S]}  \\
39 gun control  \\ 
}  \normalsize

\noindent
If you look at the full \textit{by spaces} representation for the \textit{Mayor
Blooming idiot} sentence,  you will find
10 agrees/disagrees nodes.

For our current sentence, however, we do not have sentiments nested in
sentiments.  Here is the segment of the \textit{by spaces}
display for the object of the gfbf (\textit{the rich}).  \textit{The rich}
is only in two spaces.

\small
\textsf{ \\ 
\noindent
``Republicans roared onto the post-State-of-the-Union morning shows accusing President Obama
of waging class warfare against the rich'' \\
\noindent
\noindent \textit{[1045 writer +B republicans +B obama -S]}  \\
\noindent \textit{[1047 writer -S republicans +B obama -S]}  \\
\noindent 78 the rich  \\ 
}  \normalsize

\noindent
Further, there is only a single agrees/disagrees node, the one concerning
whether Obama is waging class warfare:

\small
\textsf{ \\ 
\noindent
1029 writer disagrees with republicans that   \\
\indent 1028 isTrue   \\
\indent \indent 77 obama waging class warfare against the rich  \\ 
}  \normalsize

\noindent
Recall that two lines were crossed out in the input above.  Those
lines correspond to the Republicans having negative sentiment toward
Obama.  Here is the full set of input lines and the nodes built to
represent them:

\small
\textsf{ \\ 
\noindent
\noindent
``$\langle$alt$\rangle$ Republicans roared onto the post-State-of-the-Union morning shows   
   accusing President Obama of waging class warfare against the rich''   \\
\noindent
E1 gfbf         $\langle$obama, badFor (waging class warfare against,wagingClassWarfare:lexEntry), the rich$\rangle$  \\
\noindent B1 subjectivity $\langle$republicans, positive believesTrue (accusing), E1$\rangle$  \\
\noindent B2 privateState $\langle$writer, positive believesTrue (""), B1$\rangle$  \\
\noindent Prop1 p(B1,substantial) \\
\noindent S1 subjectivity $\langle$writer, negative sentiment (roared), republicans$\rangle$  \\
\noindent S2 subjectivity $\langle$republicans, negative sentiment (accusing), obama$\rangle$  \\
\noindent B3 privateState $\langle$writer, positive believesTrue (""), S2$\rangle$  \\ \\ 
\noindent
94 writer positive believesTrue \\
\indent    93 republicans negative sentiment  \\
\indent \indent        87 obama \\
\noindent 90 writer positive believesTrue  \\
\indent    88 republicans positive believesTrue substantial \\
\indent \indent        85 obama waging class warfare against the rich \\
\noindent 92 writer negative sentiment  \\
 \indent   89 republicans  \\ 
}  \normalsize

\noindent
For this version of the input, the system's inferences are superset
of the inferences made for the original version.
The additional inferences come from the added sentiment (node 93).
Basically, rule rule5agent infers, from node 93, a republican sentiment toward the gfbf, and
inference continues from there (as for several examples above).

Finally, here an example where the input attitude corresponding to
arguing subjectivity is believesTrue negative.

\small
\textsf{ \\ 
\noindent
\noindent
``Republicans roared onto the post-State-of-the-Union morning shows   
   denying that President Obama helped the middle class''   \\
\noindent
E1 gfbf         $\langle$obama, goodFor (helping,help:lexEntry), the middle class$\rangle$  \\
\noindent B1 subjectivity $\langle$republicans, negative believesTrue (accusing), E1$\rangle$  \\
\noindent B2 privateState $\langle$writer, positive believesTrue (""), B1$\rangle$  \\
\noindent Prop1 p(B1,substantial)  \\
\noindent S1 subjectivity $\langle$writer, negative sentiment (roared), republicans$\rangle$  \\ \\ 
\noindent
100 writer positive believesTrue   \\
\indent    98 republicans negative believesTrue substantial \\
\indent \indent        95 obama helping the middle class \\
\noindent 102 writer negative sentiment  \\
\indent    99 republicans \\ \\
\noindent
\textbf{\fbox{rule5source}}   \\
\indent
S1 sentiment toward S2 in the input \&   \\
\indent
(Assume S1 positive believesTrue) S2 privateState toward Z in input    \\
\indent    $\Longrightarrow$
S1 sentiment toward   
  S2 privateState toward Z  
\vspace{-1em}   \begin{multicols}{2}
\noindent
\textbf{Preconditions:} \\
\noindent 102 writer negative sentiment \\
\indent 99 republicans \\
\noindent
\textbf{Assumptions:} \\
\noindent 100 writer positive believesTrue \\
\indent \begin{scriptsize}98 republicans negative believesTrue substantial\end{scriptsize} \\
\indent \indent 95 obama helping the middle class \\
\vfill   \columnbreak
\noindent
\textbf{$\Longrightarrow$ Infer Node:}  \\
\noindent 
\noindent 1229 writer negative sentiment  \\
\indent \begin{scriptsize}98 republicans negative believesTrue substantial\end{scriptsize}  \\
\indent \indent 95 obama helping the middle class  \\
\noindent
\noindent 
\end{multicols}
\noindent
\textbf{\fbox{rule3.2}}   \\
\indent
S1 sentiment toward
  S2 positive/negative believesTrue substantial Z \\
\indent $\Longrightarrow$
S1 agrees/disagrees with S2 that isTrue/isFalse Z \& \\
\indent \indent S1 pos/neg believesTrue substantial Z
\vspace{-1em}   \begin{multicols}{2}
\noindent
\textbf{Preconditions:}  \\
\noindent 1229 writer negative sentiment \\
\indent \begin{scriptsize}98 republicans negative believesTrue substantial\end{scriptsize}  \\
\indent \indent 95 obama helping the middle class \\
\vfill   \columnbreak
\noindent
\textbf{$\Longrightarrow$ Infer Node:}  \\
\noindent 
\noindent 1231 writer disagrees with republicans that \\
\indent 1230 isFalse \\
\indent \indent 95 obama helping the middle class \\
\noindent
\noindent 
\noindent 1232 writer positive believesTrue substantial  \\
\indent 95 obama helping the middle class  \\
\end{multicols}  \vspace{-1em}
}  \normalsize

\noindent
The system has just inferred that the writer disagrees with the
republicans that it is false that Obama is helping the middle class
(node 1231) and, thus, that the writer believes that Obama {\it is}
helping the middle class (node 1232).

\subsection{Blocked Inferences}
\label{evidenceAgainstEGs}

Currently, we do not have general criteria for blocking
inferences toward the {\it object
of a gfbf} which is in turn the target of a sentiment.  Only ad hoc evidence
to the contrary does so.

The framework commits to one case of blocking inference
toward an {\it inanimate agent of a gfbf} which is in turn the
target of a sentiment:  when there is evidence
against the gfbf being {\it substantial} (i.e., that it happened).
Consider this example:

\small
\textsf{ \\ 
\noindent
\noindent
``Mother was worried that the tree might fall on the boy, but it didn't''   \\
\noindent
E1 gfbf         $\langle$the tree:thing, badFor (fell on,fall on:lexEntry), the boy$\rangle$  \\
\noindent S1 subjectivity $\langle$mother, negative sentiment (worried), E1$\rangle$  \\
\noindent B1 privateState $\langle$writer, positive believesTrue (""), S1$\rangle$  \\
\noindent B2 privateState $\langle$writer, negative believesTrue (""), E1$\rangle$  \\
\noindent Prop1 p(B2,substantial)  \\ \\ 
110 writer negative believesTrue substantial   \\
\indent 103 the tree fell on the boy  \\ 
\noindent
108 writer positive believesTrue   \\
\indent 106 mother negative sentiment   \\
\indent \indent 103 the tree fell on the boy   \\
\noindent
}  \normalsize

\noindent
Actually, the inference that mother is negative toward the tree is
not blocked by negative evidence; rather, the relevant rule, rule9 does not fire.  Here is rule9:

\small
\textsf{ \\ 
\noindent
\noindent
\textbf{\fbox{rule9}}   \\
\noindent
S sentiment toward A goodFor/badFor T, where A is a thing \&   \\
\noindent
(Assume S positive believesTrue substantial) A goodFor/badFor T    \\
\noindent    $\Longrightarrow$
S sentiment toward A   \\ 
}  \normalsize

\noindent
The assumption cannot be satisfied, because the only belief 
toward the gfbf is writer {\bf negative} believesTrue; for the assumption
to be made, the polarity would need to be positive.  Thus, the system
infers that mother has a positive sentiment toward the boy, but not
that she has a negative sentiment toward the tree (see Appendix A).

There are two types of evidence we hypothesize systematically block
inferences toward an {\it animate agent of a gfbf} which is in turn
the target of a sentiment:  evidence against the
gfbf being intentional (e.g., not an accident), and evidence
against the agent being positive toward the idea of the gfbf.
Either type of evidence blocks the reasoning chain illustrated in
Section \ref{animateAgent}, namely the chain from gfbf, to intention toward the
gfbf, toward positive sentiment toward the idea of the gfbf, to
agreement/disagreement with the agent, to a positive/negative attitude
toward the agent.

Consider the following example, which we looked at above in
Section \ref{compositionality}:

\small
\textsf{\\
\noindent
\noindent
``Oh no! The tech staff tried to stop the virus, but they failed.''   \\
\noindent
E1 gfbf         $\langle$the tech staff, badFor (stop,stop:lexEntry), the virus (virus:lexEntry)$\rangle$  \\
\noindent I1 influencer   $\langle$the tech staff, reverse (failed,fail:lexEntry), E1$\rangle$  \\
\noindent S1 subjectivity $\langle$writer, negative sentiment (Oh no!),I1$\rangle$  \\
\noindent V1 evidence     $\langle$none,positive intends (tried),E1$\rangle$  \\
\noindent V2 evidence     $\langle$none,negative believesTrue (failed),E1$\rangle$  \\ \\ 
\noindent
\noindent  115 writer negative sentiment \\
\indent   114 the tech staff $\langle$reverse$\rangle$ \\
\indent \indent   111 the tech staff stop the virus \\
\noindent
118 There is evidence that the following is not substantial \\
\indent    111 the tech staff stop the virus  \\
\noindent 117 There is evidence that the following is intentional:  \\
\indent    111 the tech staff stop the virus  \\
\noindent
\noindent
\textbf{influencer} node: \\
 \noindent \textit{115 writer negative sentiment}   \\
\indent   114 the tech staff $\langle$reverse$\rangle$  \\
\indent \indent   111 the tech staff stop the virus \\
\noindent
\textbf{new gfbf} node:  \\
\noindent  \textit{1262 writer negative sentiment}   \\
\indent   1259 the tech staff goodFor the virus  \\
\noindent
\textbf{new evidence} nodes:   \\
\noindent
\noindent 1260 There is evidence that the following is not intentional:  \\
\indent 1259 the tech staff goodFor the virus   \\
\noindent 1261 There is evidence that the following is substantial \\
\indent 1259 the tech staff goodFor the virus \\ 
}  \normalsize 

\noindent
Recall from Section \ref{compositionality} that \textit{failure} is a {\it reverser} influencer, resulting
in new gfbf and evidence nodes, as just shown.\\

The system infers that the writer is negative toward the virus,
but not that he is negative toward the tech staff.  Specifically, rule6 is
blocked from firing:

\small
\textsf{ \\ 
\noindent
\noindent
\textbf{\fbox{rule6}}   \\
\indent A goodFor/badFor T, where A is animate    \\
\noindent $\Longrightarrow$
A intended A goodFor/badFor T   
\vspace{-1em} \begin{multicols}{2}
\noindent
 \textbf{Preconditions:} \\
\noindent  \textit{1262 writer negative sentiment}  \\
\indent   1259 the tech staff goodFor the virus 
\vfill    \columnbreak
\noindent
\textbf{$\Longrightarrow$ Infer Node:} \\
Blocked by evidence node  \\
\noindent The evidence node:   \\
\noindent 1260 ... the following is not intentional:  \\
\indent 1259 the tech staff goodFor the virus   \\ 
\end{multicols}    \vspace{-1em}
}  \normalsize

\noindent
The following sentence illustrates an inference being blocked because the
agent is not positive toward their intentional event.

\small
\textsf{ \\ 
\noindent
\noindent
``Thanks to the Affordable Care Act, consumers will receive more value  
for their premium dollar because insurance companies will be required to spend  
80 to 85 percent of premium dollars on medical care and health care 
quality improvement, rather than on administrative costs, starting in 2011.''   \\
\noindent
E1 gfbf         $\langle$insurance-companies, goodFor (spend on,spendOn:lexEntry), health-care-quality-improvement$\rangle$  \\
\noindent B1 privateState $\langle$writer, positive believesTrue (""), E1$\rangle$  \\
\noindent S1 subjectivity $\langle$writer, positive sentiment (Thanks \& value), E1$\rangle$  \\
\noindent V1 evidence     $\langle$insurance-companies, negative sentiment (required), E1$\rangle$  \\  \\
\noindent
122 writer positive believesTrue   \\
\indent    119 insurance-companies spend on health-care-quality-improvement  \\
\noindent 124 writer positive sentiment  \\
\indent    119 insurance-companies spend on health-care-quality-improvement  \\
\noindent 125 (evidence,insurance-companies,negative,sentiment) \\
\indent   126 ideaOf \\
\indent \indent        119 insurance-companies spend on health-care-quality-improvement \\ 
}  \normalsize

\noindent
The input includes one gfbf, node 119; sentiment on the part of the
writer, node 124 (\textit{Thanks to}, \textit{value}); and evidence against the agent
being positive toward the event.  The gfbf is intentional -- it
will not be an accident that the insurance companies will spend their
premium dollars -- but, they will be {\it forced to} (and won't like
it).  The evidence node blocks rule7 from firing, and, thus, the system does
{\it not} infer that the writer is positive toward the insurance
companies, even though it does infer the writer is positive toward
health-care-quality-improvement, and the insurance companies are doing
something goodFor it  (see Appendix A).\\

\section{Exploring Interdependent Ambiguities}
\label{exploring}
Recall this example from Section \ref{sentobject}:

\small
\textsf{ \\ 
\noindent
\noindent
"Is it no surprise then that MoveOn would attack Senator McCain.!?"   \\
\noindent
4 writer negative sentiment  \\
\indent 1 MoveOn attack Senator McCain \\ 
}  \normalsize 

Note that the subjectivity clues in this sentence --  fronting, {\it
surprise}, and {\it then} -- are ambiguous
with respect to whether the
writer is expressing a positive or negative sentiment.  If the
writer were positive rather than negative toward MoveOn attacking
McCain, then the polarities of the inferred sentiments would be
reversed.  Considering all the rules together, we really only have two
choices rather than eight:
{\bf (1)} polarity(\textit{surprise}) = negative; sent(writer,McCain) = positive;
sent(writer,MoveOn) = negative), or {\bf (2)} polarity(\textit{surprise}) = positive, sent(writer,McCain) = negative;
sent(writer,MoveOn) = positive.
From an NLP perspective, the interdependencies captured by the
implicature rules may be encoded as constraints to support sentiment
propagation among entities (\cite{dengwiebeeacl2014}) and as constraints
in an optimization framework for joint disambiguation.\footnote{In submission.}

Further, this example illustrates that evidence from the larger discourse or
pragmatic situation could be exploited to improve sentence-level processing.
Any outside evidence (for example, that the writer is politically
liberal or conservative) concerning the writer's attitude toward
McCain or MoveOn could be used to determine which {\bf set} of
attitudes is more probable.

\section{A Return to the KR Scheme: S negative believesTrue
  T}
\label{negBelMeaning}
This section may safely be skipped.\\

We stated above in Section \ref{kr} that
S {\bf negative} believesTrue T means
(A) or (B):
\begin{itemize}
\item[(A)]
S does not believe that T, in the sense that T is not in S's belief space.
Among the things that S believes are true, T is not one of them.
S may not believe anything about T.
\item[(B)]
S does not believe that T, in the sense that S believes that not T, i.e., S believes that T
is false.
\end{itemize}

Consider this sentence given in Section \ref{evidenceAgainstEGs}:

\small
\textsf{ \\ 
\noindent
\noindent
"Mother was worried that the tree might fall on the boy, but it didn't"   \\
\noindent
E1 gfbf  $\langle$the tree:thing, badFor (fell on,fall on:lexEntry), the boy$\rangle$ \\
\noindent S1 subjectivity $\langle$mother, negative sentiment (worried), E1$\rangle$ \\
\noindent B1 privateState $\langle$writer, positive believesTrue (""), S1$\rangle$ \\
\noindent B2 privateState $\langle$writer, negative believesTrue (""), E1$\rangle$ \\
\noindent Prop1 p(B2,substantial) \\ 
}  \normalsize

In this sentence, the tree did not in fact fall on the boy.  Thus,
we have case (B).  The input line corresponding to the tree not
falling is B2.  This line
means either (A) the writer has an absence of belief that the tree
fell on the boy, or (B) the writer believes that it is false
that the tree fell on the boy.  Our KR scheme does not ``nail
down'' which of (A) or (B) applies to this sentence, even though a
human can perceive that it is (B).

Contrast this with the following (which is included in the appendix output file):

\small
\textsf{ \\ 
\noindent
\noindent
"Mother dislikes the judge; by the way, he freed the  
murderer, but Mother doesn't know he did."   \\
\noindent
\noindent S1 subjectivity $\langle$Mother, negative sentiment (dislikes), the judge$\rangle$ \\
\noindent B1 privateState $\langle$writer, positive believesTrue (""), S1$\rangle$ \\
\noindent E1 gfbf\indent \indent  $\langle$the judge, goodFor (freeing,free:lexEntry), the murderer$\rangle$ \\
\noindent B2 privateState $\langle$writer, positive believesTrue (""), E1$\rangle$ \\
\noindent B3 privateState $\langle$mother, negative believesTrue (doesn't know), E1$\rangle$ \\
\noindent B4 privateState $\langle$writer, positive believesTrue (""), B3$\rangle$ \\ 
}  \normalsize

The relevant private state is B3, that mother negative
believesTrue that the judge freed the murderer.  This sentence
illustrates case (A):  we only know she has the absence of belief that
the judge freed the murderer.  The sentence does not suggest that
mother believes that it is false that the judge freed the murderer.
Again, our KR scheme does not commit to which of (A) or (B) applies,
even though a human can perceive that it is (A).

Consider the arguing subjectivity examples from Section
\ref{arguingSubjectivity}.
The positive arguing true example we gave is this:

\small
\textsf{ \\ 
\noindent
\noindent
"Republicans roared onto the post-State-of-the-Union   
\noindent
morning shows accusing President Obama of waging class warfare against the rich"  \\ 
}  \normalsize

The relevant private state is that the Republicans positive believeTrue
that Obama badFor the rich.  The system makes several
inferences, leading to these:

\small
\textsf{ \\ 
\noindent
1029 writer disagrees with republicans that \\
\indent 1028 isTrue  \\
\indent \indent 77 obama waging class warfare against the rich  \\
\noindent 1030 writer negative believesTrue substantial  \\
\indent 77 obama waging class warfare against the rich \\ 
}  \normalsize

Together, 1029 and 1030 demonstrate a (conceptual) weakness of our KR scheme combined
with this inference rule. Node 1029 implies that the writer believes it
is false that obama badFor the rich (since the writer disagrees with
the republicans that it is true).   That is, 1029 suggests that case (B)
applies.  But, node 1030 loses that information:  all we know from 1030 is
that case (A) or case (B) applies. 

We do not anticipate this being a problem for our project as it
proceeds, as we are not aiming to make such fine-grained
belief distinctions.  If it does become a problem, we can refine the KR
scheme.  This would require adding logical negation (and rules to
reason with it) to the rule-based framework, which would
increase its overall complexity.

Note that we {\it are} incorporating negation into
the framework via the compositionality component (see Section
\ref{compositionality}).  Compositional processing is performed {\it prior} to the inference
process.  An advantage of this is that the two processes - semantic
compositionality and pragmatic inference - do not need to share the
same representations or mechanisms.

\section{A Return to Space Extension: Extensions that could
  potentially be added}
\label{extensionsThatCouldBeAdded}
This section may safely be skipped.

As described in Section \ref{attitudeAscription}, when a rule fires,
the As and Qs are
placed into all the spaces that already contain all the Ps.  Then, the
Ps, As, and Q's are added to the spaces created by replacing
sentiments with beliefs on those paths.    The idea is, if S has a
sentiment toward T, then S must have a positive believesTrue attitude
toward T.  In this section, we will call the new spaces the {\it expected spaces}.

There are cases where we opt not to attribute beliefs, when
arguably it would be valid to attribute them.    These cases are
beliefs toward gfbfs, ideaOfs, animate entities, and things.

Consider these two segments from a \textit{by spaces} display:

\small
\textsf{ \\ 
\noindent
\noindent
Mother is upset that the judge freed the murderer.   \\
\noindent
\textit{[215 writer +B mother -S]}   \\
\noindent
\textit{[216 writer +B mother +B]}   \\
\noindent
213 judge positive intends   \\
\indent 5 judge freeing the murderer   \\
\noindent
\textit{[212 writer +B mother -S]}   \\
\noindent
\textit{[225 writer +B mother -S judge +S]}   \\
\noindent
\textit{[227 writer +B mother +B judge +S]}   \\
\noindent
6 the murderer  \\ 
}  \normalsize

Node 216 is the result of adding a conclusion to one of the {\it
expected} spaces, i.e., the result of replacing -S in the path of 215
with +B.   Note that such nodes are excluded for node 6:  the
rightmost node of all the spaces in which 6 appears is a sentiment.

The difference is that 213 is a private state, and 6 is an entity.
private states are propositions; they naturally may be the objects of {\it
believes that}.  While we have a means for representing {\it believes
that} having an object that is not a proposition (we specify that the source
believes that p(object) for some p) it seems more natural not to
introduce them without a reason to.

The case of {\it believes that} objects that are not propositions which
currently arises is when they are added to the input, for example,
node 87 in the following:

\small
\textsf{ \\ 
\noindent
\noindent
"""Mother dislikes the judge for freeing the murderer.   \\
\noindent
88 writer positive believesTrue   \\
\indent 83 Mother negative sentiment   \\
\indent \indent 84 the judge   \\
\noindent
87 writer positive believesTrue   \\
\indent 86 the judge freeing the murderer  \\ 
}  \normalsize

We wanted to allow inputs such as 87; we tried some other more complex
input schemes, but the reactions to them suggested we should stick to
the simpler scheme.  

Now, perhaps all possible beliefs should be generated.  After all, if
you have a sentiment toward something you have to believe {\it something}
about it.  In that case, the entry for 6 would be the following (the
starred nodes would be the additional one).

\small
\textsf{ \\ 
\noindent
\noindent
Mother is upset that the judge freed the murderer.   \\
\noindent
[212 writer +B mother -S]   \\
\noindent
[225 writer +B mother -S judge +S]   \\
\noindent
[227 writer +B mother +B judge +S]   \\
\noindent
[* writer +B mother +B]   \\
\noindent
[* writer +B mother +B judge +B]   \\
\noindent
6 the murderer  \\ 
}  \normalsize

This would be a straightforward change to make.

\section{Related Work}
\label{relatedwork}
In this section, we first discuss recent related work on sentiment
analysis in NLP, then acknowledge older work in NLP and AI whose ideas
we exploited to create the inference architecture.

\subsection{Sentiment Analysis}

While most work in NLP addresses explicit sentiment, there is work
that addresses implicit sentiment.  Though none presents, as we do,
general implicature rules relating gfbf events and explicit and
implicit sentiments ascribed to sources/holders, several previous
works address relevant aspects of our overall framework.  By design,
our framework abstracts away from specific linguistic realizations, to
capture general underlying inference patterns (for example, gfbf spans
may be verbs as well as nouns, and gfbf polarity reversal may involve
multiple compositions involving words of different parts of speech).
Many previous papers address relevant topics in more linguistic
depth than we do.  Our hope is that their results may be exploited in
the future to realize fully automatic framework components, and that
our work will help integrate their various findings.

Researchers have investigated identifying objective words that have
positive or negative connotations (e.g., \cite{fengacl2013}) and
identifying noun product features that imply opinions (e.g.,
\cite{popescuetzioni2005,zhang-liu:2011:ACL-HLT2011}).  Rule 10 in our
schema shows where connotation is brought into the framework: in the
absence of evidence to the contrary, Rule 10 infers sentiment from
connotation.  The simple lexicon test for connotation in the
precondition of Rule 10 could be replaced by a more sophisticated
recognition component that handles related notions such as polar
features.

Several papers apply compositional semantics to determine
polarity 
(e.g.,
\cite{Moilanen2007,choi-cardie:2008:EMNLP,moilanen-pulman:2009:RANLP09,moilanen2010packed,neviarouskayaetalNLE2011,ruppenhofer-rehbein:2012:WASSA2012}). 
The goal of such work is to
determine one overall polarity of an expression or sentence (though
polarities may be assigned to intermediate entities along the way).
Conceptually, in terms of the framework, such composition is performed
before inference begins. 
(Section \ref{compositionality} describes
the compositionality currently performed by the system, namely
processing influencer chains ending in gfbf events.  Compositional
processing should also be incorporated into its recognition of
connotation for Rule 10, building on the previous work just cited.)
In contrast to compositional processing, the
implicature rules infer sentiments of 
{\it different} sources/holders toward 
various events and entities in the sentence.  In addition, their inferences are defeasible.  

States and events (such as gfbf events) which positively
or negatively affect entities have figured in several works in
sentiment analysis.   For example, two papers mentioned above
\cite{zhang-liu:2011:ACL-HLT2011,choi-cardie:2008:EMNLP}
include linguistic patterns for the tasks that they address that
include gfbf events.
\cite{COIN:COIN455}, in their work toward automating plot units, 
generate a lexicon of \textit{patient
polarity verbs}, which correspond to gfbf events whose spans are
verbs.  \cite{riloff-EtAl:2013:EMNLP}, in their
work on recognizing sarcasm on Twitter, learn phrases describing
situations which are negative for the Tweeter.  

Turning to the {\it inference} of implicit from explicit sentiment,
\cite{zhang-liu:2011:ACL-HLT2011} introduce two specific rules:
$Decreased Neg \rightarrow Positive$ and $Decreased Pos \rightarrow
Negative$ (p. 577), which apply to phrases such as \textit{reduce the
fun of driving}.  \cite{COIN:COIN455} infer the affective states of
characters in fables, for example, that Mary has positive affect from
\textit{Mary laughed} and that John has positive affect from
\textit{John was rewarded}.  These are different inferences from the
ones we address: we do not infer the affective states {\it of}
entities mentioned in the text, but rather sentiments {\it held by}
the writer and other entities {\it toward} the entities and events in the
sentence.  Though \cite{riloff-EtAl:2013:EMNLP} do not perform
inference, per se, they do address contrasts between polarities of
explicit and implicit sentiments as we do -- in their model, sarcasm
arises from positive sentiment toward negative situations.

In Section \ref{blocking}, we saw that inferences toward agents may be
blocked by evidence breaking an inference chain such as \textit{Agent
gfbf Object} $\rightarrow$ \textit{Agent intended Agent gfbf Object}
$\rightarrow$ \textit{Agent positive sentiment toward ideaOf Agent
gfbf Object}.  These inference chains were partially inspired by
the study of implicit sentiment by \cite{Greene2009}.
They investigated a ``connection between implicit
sentiment and grammatically relevant semantic properties $\dots$ by
varying the syntactic form of event descriptions'' (p. 505) and show
that the semantic properties of descriptions predict perceived
sentiment.  They constructed stimuli of the form 
\textit{X verb of killing Y} in the context of news reports
about crimes, where \textit{Y} is a victim.  Since killing a victim is
rarely viewed positively, their stimuli correspond, in our terms, to: \\

\noindent
$E_1$: $\langle$X,kill:{\sc badFor}, Y$\rangle$\\
$S_1$:  sent(writer,$E_1$) = negative\\ 

They varied the syntactic form in
ways corresponding to
semantic properties. 
Subjects were asked to rate how sympathetic they
perceive a stimulus to be toward the agent, $X$.  They found 
that volition is negatively correlated with sympathy -- the more
volitional the act, the more negative the judgment against $X$.  
Consistent with their findings, in our system, evidence
against $X's$ action being volitional blocks a negative inference toward 
$X$.  Their work will be relevant in future work on detecting
linguistic evidence against agent inferences, i.e., detecting evidence for defeated
implicatures.

\subsection{Inference Architecture}

Our inference architecture involves explicit rules and mechanisms
for default inference and inference within private-state spaces.

The assumptions in the rules are inspired by
\cite{Hobbs:1993:IA:162229.162232}, where interpreting a text is cast
in the form of abduction, and assumptions are made as necessary to
derive an interpretation.
All assumptions in our rules are assumptions of private states, either
beliefs or sentiments.
Our ascription of belief was inspired by work of Wilks and colleagues
(e.g., \cite{WilksBien:1983,Wilks:1987:MAH:1625015.1625038}) who
devise a \textit{default ascriptional rule} which assumes
``one's view of another person's view is the same as one's own
\textit{except where there is explicit evidence to the contrary}''
(\cite{Wilks:1987:MAH:1625015.1625038}(p. 119)).

Our main inspiration for default inference is work done in the 
1970's by Schank and his research group.  The inference performed by
our system is forward inference, triggered by the input, resulting in
reasoning chains that proceed to the end unless evidence to the
contrary breaks the chain (halts inference).  This is similar to the type of
reasoning performed by SAM (Script Applier Mechanism)
(\cite{schank1977}), for example (though our rules do not instantiate
roles along the way).

Our main inspiration for inference within private-state spaces is work
by Stuart Shapiro, William Rapaport, and their research
group (\cite{Rapaport86,ShapiroRapaport87}) on their knowledge
representation system, SNePS, considered as a fully intensional
propositional semantic network, and the extension of SNePS, created by
Jo\~ao Martins, to handle reasoning in multiple belief spaces

\section{Conclusions}
\label{conclusions}
While previous sentiment analysis research has concentrated on the
interpretation of explicitly stated opinions and attitudes, this work
initiates the computational study of a type of opinion implicature
(i.e., opinion-oriented inference) in text.   This paper
described a rule-based framework for representating and analyzing
opinion implicatures which we hope will contribute to deeper automatic
interpretation of subjective language.  In the course understanding implicatures,
the system recognizes implicit sentiments (and beliefs)
toward various events and entities in the sentence, often attributed
to different sources and of mixed
polarities; thus, it produces a much richer interpretation than is
typical in opinion analysis.

\bibliographystyle{spbasic}      
\bibliography{jans,other,master,subjectivity,lingjia}   

\end{document}